%% file: main.tex
\documentclass[10pt,twocolumn,letterpaper]{article}
\usepackage[pagenumbers]{arxiv}
\input{preamble}

\definecolor{linkblue}{rgb}{0.21,0.49,0.74}
\usepackage[pagebackref,breaklinks,colorlinks,allcolors=linkblue]{hyperref}

\definecolor{titleorange}{rgb}{0.85,0.45,0.05}
\newcommand{\hlsteer}{\texorpdfstring{\textbf{\textcolor{linkblue}{Steer}}}{Steer}}
\newcommand{\hlone}{\texorpdfstring{\textbf{\textcolor{titleorange}{One}}}{One}}
\title{\method: Unlocking \hlsteer able General Humanoid Motion Tracking\\ via \hlone-Step Diffusion Whole-Body Control}

\author{
  Yifei Huang$^{1,*}$ \quad
  Yazhao Yang$^{2,*}$ \quad
  Weijia Li$^{3}$ \quad
  Tianle Shi$^{4}$ \quad
  Ji Jin$^{5}$ \\
  Zhihui Cai$^{2}$ \quad
  Zihao Liu$^{6,7}$ \quad
  Jianzhuang Zhao$^{6}$ \quad
  Songlin Dong$^{5}$ \quad
  Xinyuan Chang$^{7}$ \\
  Yihong Gong$^{8}$ \quad
  Dian Shen$^{1,\dagger}$ \quad
  Zhiheng Ma$^{5,7,\dagger}$ \\[0.4em]
  $^{1}$Southeast University \quad
  $^{2}$Southern University of Science and Technology \\
  $^{3}$The Hong Kong Polytechnic University \quad
  $^{4}$Shenzhen University \\
  $^{5}$Shenzhen University of Advanced Technology \quad
  $^{6}$Northwestern Polytechnical University \\
  $^{7}$Primotion \quad
  $^{8}$Xi'an Jiaotong University \\[0.3em]
  {\tt\small \{230260231, dshen\}@seu.edu.cn} \quad
  {\tt\small 12533265@mail.sustech.edu.cn} \\
  {\tt\small alistair.li@connect.polyu.hk} \quad
  {\tt\small mazhiheng@suat-sz.edu.cn} \quad
  {\tt\small \{zl, xc, zm\}@primotion.com} \\[0.3em]
  {\small $^*$Equal contribution \quad $^\dagger$Corresponding authors} \\[0.1em]
  {\small Project page: \url{https://onesteer.github.io}}
}

\begin{document}
\twocolumn[{%
\renewcommand\twocolumn[1][]{#1}%
\maketitle
\input{fig/teaser}
}]
\input{sec/0_abstract_new}
\input{sec/1_intro_new}
\input{sec/2_related}
\input{sec/3_method_new}

\input{sec/4_exp}
\input{sec/5_conclusion}

{
  \small
  \bibliographystyle{ieeenat_fullname}
  \bibliography{references}
}
\clearpage
\appendix
\renewcommand{\thesection}{S\arabic{section}}
\renewcommand{\thetable}{S\arabic{table}}
\renewcommand{\thefigure}{S\arabic{figure}}
\renewcommand{\theequation}{S\arabic{equation}}
\renewcommand{\theHsection}{S\arabic{section}}
\renewcommand{\theHtable}{S\arabic{table}}
\renewcommand{\theHfigure}{S\arabic{figure}}
\renewcommand{\theHequation}{S\arabic{equation}}
\setcounter{table}{0}\setcounter{figure}{0}\setcounter{equation}{0}
\input{sec/6_appendix}
\end{document}

%% file: preamble.tex
\usepackage{cuted} 

\usepackage{currfile} 

\usepackage{caption} 
\usepackage{microtype}

\makeatletter
\renewcommand\paragraph{\@startsection{paragraph}{4}{\z@}{2pt}{-0.5em}{\normalfont\normalsize\bfseries}}
\makeatother

\usepackage{booktabs}
\usepackage{array}
\usepackage{multirow}
\usepackage{graphicx}
\usepackage{amsmath,amssymb}
\usepackage{xcolor}
\usepackage{enumitem}
\newcommand{\method}{OneSteer}
\newcommand{\supp}[1]{\cref{#1} in the appendix}  

%% file: fig/teaser.tex
\begin{center}
  \centering
  \vspace{-2mm}
  \includegraphics[width=0.98\textwidth]{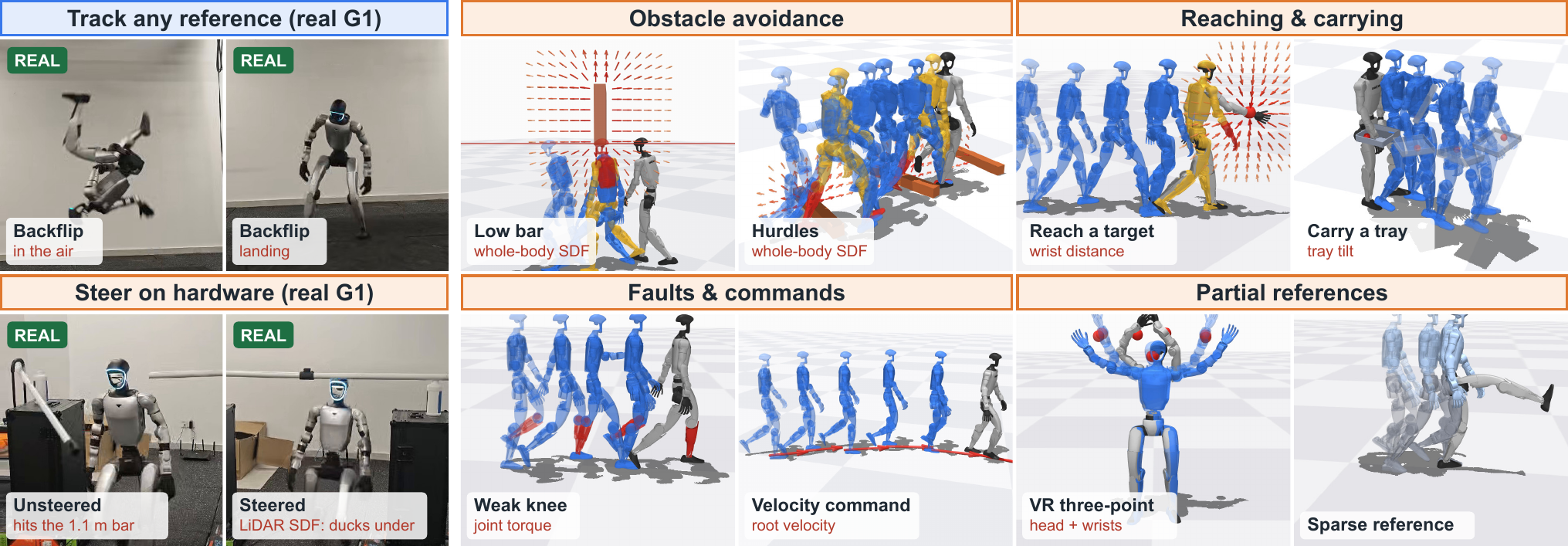}
  \vspace{-1mm}
  \captionof{figure}{\textbf{One frozen general tracker, steerable at test time.} Left: on a Unitree G1, \method{} tracks diverse references, here a backflip (top); with an SDF cost from the onboard LiDAR, it passes under a 1.1\,m bar that the unsteered controller hits (bottom). Right: with identical frozen weights, differentiable trajectory costs or body-group masks adapt behavior zero-shot without retraining across diverse simulation tasks (blue: earlier executed poses, the latest one in the robot's own materials; gold: the reference pose at the critical moment).}
  \label{fig:teaser}
  \vspace{-2mm}
\end{center}

%% file: sec/0_abstract_new.tex
\begin{abstract}
Deployment objectives often specify desired outcomes
without providing revised trajectories, while parts
of the original reference must still be followed.
We present \method{}, a one-step diffusion whole-body
controller that combines general motion tracking
with selective test-time steering.
To accommodate new objectives, differentiable task costs
guide optimization of the generator's input noise
with frozen weights, while body-group reference masks
relax selected tracking requirements.
Predicted state--action plans provide a shared interface
for task optimization and feedback-conditioned execution.
For nominal tracking, the model trained from scratch
achieves the lowest local tracking error among the
evaluated baselines on successful held-out AMASS and
BONES clips.
For steering, one frozen distilled controller adapts
across diverse deployment objectives in simulation,
raising the clean passage rate under a 1.25\,m bar
from 2\% to 90\%.
On a physical Unitree G1, \method{} completes all 27
nominal tracking trials across nine diverse whole-body
motion sequences without falling.
With a steering cost derived from a signed distance field
built using only onboard LiDAR, it also passes under a
1.1\,m low bar in four of five attempts and steps over a
12\,cm hurdle without falling in all three attempts.
Beyond control, prediction variation in the model trained
from scratch ranks likely tracking failures without
physics rollouts.
This scoring method is more than $10\times$ faster than
simulation-based scoring with five rollouts per clip
and guides targeted fine-tuning.
Project page: \url{https://onesteer.github.io}.
\end{abstract}

%% file: sec/1_intro_new.tex
\section{Introduction}
\label{sec:intro}

General motion tracking enables a single humanoid policy
to follow diverse reference motions
~\cite{sonic,mimiclite,gmt,hover,omnih2o}.
References encode detailed timing and whole-body
coordination, but deployment often imposes requirements
beyond faithful reproduction.
A robot may need to clear obstacles, stabilize a tray,
or operate under reduced torque limits.
These requirements specify desired outcomes without
supplying revised whole-body trajectories.
Meeting them calls for adapting the motion while
reusing the coordination learned through general tracking.

Editing the reference with an upstream planner offers
one route to steering~\cite{closd,textop,omg}, but the
resulting trajectory must still be executed by a separate
tracker, which can introduce a gap between planned
and executed motion~\cite{predactor}.
Guided generative controllers offer another route by
steering learned behavior priors with task
costs~\cite{diffuseclock,beyondmimic,uniphys,predactor}.
However, these controllers are conditioned on goals,
keyframes, text, or a trajectory rather than on a
full-body reference window.
For general tracking, the controller must adapt to
the new objective while continuing to follow the
reference components that have not been relaxed.
Since both the prescribed motion and the deployment
objective can vary, this raises a central question:
\textit{Can a single controller reuse its tracking competence
across new motion--objective combinations?}

This calls for a reference-conditioned controller
that supports efficient test-time steering.
We introduce \method{} (\cref{fig:teaser}), a whole-body
controller built around a \textbf{one-step} diffusion generator.
Conditioned on proprioception and a future reference window,
the generator maps input noise directly to a short
state--action plan in one forward pass.
This supports low-latency nominal tracking and allows
each task-guidance update to differentiate through
a single generator evaluation, without unrolling
an iterative denoising chain.

For new objectives, we optimize the generator's input noise
using differentiable task costs while keeping
network weights fixed.
Body-group reference masks relax selected tracking
requirements, while unmasked components remain
tracking targets.
An inverse-dynamics head combines the optimized
near-term states with current robot feedback to
produce commands, followed by replanning at the
next control step.
The state plan thus serves as a shared interface
for task optimization and feedback-conditioned execution,
allowing new objectives to reuse the same tracking controller.

The predictive representation also provides an empirical
signal of tracking difficulty.
With proprioception and the reference held fixed,
we measure variation across plans generated from
different noise inputs and use it to rank likely
tracking failures without physics rollouts for scoring.
The resulting ranking guides targeted fine-tuning.

We train the controller either by distillation from
an RL teacher or from scratch using model-based RL,
and evaluate tracking on held-out AMASS and BONES motions.
On successful clips, the model trained from scratch
achieves the lowest local tracking error among the
evaluated baselines on both benchmarks:
26.5\,mm on AMASS and 22.6\,mm on BONES.
Nominal inference for the distilled controller takes
3.8\,ms per generator call on an RTX 4090 GPU.
On a physical Unitree G1, \method{} completes all 27
nominal tracking trials across nine diverse whole-body
motion sequences without falling.
With a steering cost derived from a signed distance field
built using only onboard LiDAR, it also passes under a
1.1\,m low bar in four of five attempts and steps over a
12\,cm hurdle without falling in all three attempts.
In simulation, one frozen distilled controller adapts
across obstacle clearance, reaching, reduced torque limits,
tray stabilization, and velocity commands.
Across 79 walking motions from unseen actors, guidance
increases the clean passage rate under a 1.25\,m bar
from 2\% to 90\%.
For failure assessment, prediction variation on a model
trained from scratch attains 90\% precision among the
highest-ranked 2\% of 7{,}884 reference clips.
Scoring requires no physics rollouts and is more than
$10\times$ faster than simulation-based scoring with
five rollouts per clip.

Our contributions are:
\begin{itemize}
    \item \textbf{One-step predictive general tracking.}
    We develop a reference-conditioned whole-body controller
    that generates state--action plans in one forward pass,
    enabling low-latency tracking and differentiable
    test-time refinement.
    The distilled controller achieves 3.8\,ms nominal
    inference on an RTX 4090 GPU.

    \item \textbf{Selective steering with frozen weights.}
    We combine cost-guided input optimization, body-group
    reference relaxation, and feedback-conditioned execution
    to adapt the same tracking controller to new objectives
    while retaining unmasked tracking targets.
    In simulation, one frozen distilled controller raises
    the clean passage rate under a 1.25\,m bar from
    2\% to 90\% across 79 walking motions from unseen actors.

    \item \textbf{Failure discovery from prediction variation.}
    We introduce an empirical score based on variation
    across generated plans to rank likely tracking failures
    and guide targeted fine-tuning.
    On the model trained from scratch, it attains 90\%
    precision among the highest-ranked 2\% of 7{,}884
    reference clips, with scoring requiring no physics
    rollouts and running more than $10\times$ faster
    than simulation-based scoring with five rollouts
    per clip.
\end{itemize}

%% file: sec/2_related.tex
\section{Related Work}
\label{sec:related}

\paragraph{Humanoid motion tracking.}
Physics-based imitation~\cite{deepmimic} has scaled to general humanoid trackers that follow arbitrary reference motions.
SONIC~\cite{sonic} trains on hundreds of hours of motion capture with unified tokens, BFM-Scale~\cite{bfmscale} treats tracking as pre-training for a behavior foundation model, MimicLite~\cite{mimiclite} provides an efficient training and deployment recipe, and GMT~\cite{gmt}, HOVER~\cite{hover}, OmniH2O~\cite{omnih2o}, UniTracker~\cite{unitracker}, RGMT~\cite{rgmt}, OmniTrack~\cite{omnitrack} and MOSAIC~\cite{mosaic} improve robustness and coverage.
These controllers are deterministic policies whose behavior is fixed by the reference.
Several works add flexibility on top of tracking.
Latent skill models such as ASE~\cite{ase}, CALM~\cite{calm} and PULSE~\cite{pulse} reuse a learned latent space but train a new high-level policy per task.
BFM-Zero~\cite{bfmzero} embeds motions, goals and rewards in a shared latent space with forward-backward representations, which makes the policy promptable but requires careful RL training.
Perceptive BFM~\cite{perceptivebfm} and ViBe~\cite{vibe} add vision or RL adaptation, at the cost of terrain-specific data or retraining when the task changes, and ConstrainedMimic~\cite{constrainedmimic} wraps an RL tracker with operational-space barrier functions.
In our work, the tracker itself is generative, so any differentiable cost can be applied at test time without retraining.

\paragraph{Generative whole-body control.}
Generative models~\cite{diffuser,decisiondiffuser,pdp} enter whole-body control in two ways.
Hierarchical pipelines such as CLoSD~\cite{closd}, TextOp~\cite{textop}, UniAct~\cite{uniact}, OMG~\cite{omg}, TEXEDO~\cite{texedo} and Trace and Pace~\cite{tracepace} use a generative planner to produce or edit kinematic references~\cite{motionduet} for a separate low-level tracker.
Even with closed-loop replanning, the planner does not know what the tracker can execute, and the resulting planner--controller mismatch~\cite{predactor} leads to infeasible motions and collisions.
Integrated controllers such as BeyondMimic~\cite{beyondmimic}, Diffuse-CLoC~\cite{diffuseclock}, PredActor~\cite{predactor}, UniPhys~\cite{uniphys}, SCDP~\cite{scdp} and ADAPT~\cite{adapt} instead model states and actions jointly (\cref{tab:related}), but they are conditioned on sparse goals, keyframes or text rather than full-body reference horizons, and therefore cover the tasks seen in training rather than general tracking.
Concurrently, OmniXtreme~\cite{omnixtreme} tracks highly dynamic skills with a multi-step flow policy but does not support test-time steering.
\method{} is an integrated controller conditioned on a sliding full-body reference window: it tracks general motions in one step, and its plans can be steered by costs in the closed loop.

\begin{table}[t]
\centering
\small
\setlength{\tabcolsep}{3pt}
\resizebox{\linewidth}{!}{%
\begin{tabular}{@{}lcccccc@{}}
\toprule
 & Body & Real & Trained on & Conditioning & Tracking & Steer \\
\midrule
BeyondMimic~\cite{beyondmimic} & G1 & yes & VAE rollouts & goals, keyframes & no & yes \\
Diffuse-CLoC~\cite{diffuseclock} & char. & no & tracker rollouts & keyframes & no & yes \\
PredActor~\cite{predactor} & G1 & yes & tracker rollouts & text & no & yes \\
UniPhys~\cite{uniphys} & char. & no & tracker rollouts & text, trajectory, goal & no & yes \\
SCDP~\cite{scdp} & G1 & yes & expert rollouts & velocity or ref.\ frame & walking & no \\
\midrule
\method{} (ours) & G1 & yes & tracker / RL & reference window & \textbf{general} & \textbf{yes} \\
\bottomrule
\end{tabular}}
\caption{Diffusion controllers predicting joint state-action trajectories. Real: deployed on a robot; char.: simulated SMPL character; Conditioning: test-time input besides the robot state; Tracking: follows arbitrary full-body references; Steer: test-time cost guidance.}
\label{tab:related}
\end{table}

\paragraph{One-step generation and training-free guidance.}
The latency of iterative diffusion~\cite{diffusionpolicy} can be removed by distillation~\cite{onedp,ofp,consistencypolicy,manicm}, by flow-based objectives that admit one-step sampling~\cite{rectifiedflow,meanflow}, or by training one-step generators directly, as in drifting models~\cite{drifting} and priming models~\cite{priming}, which first train the generator on a denoising reconstruction objective.
In legged locomotion, RFPO~\cite{rfpo} distills an RL policy into a one-step rectified-flow policy, without reference tracking or steering.
To our knowledge, \method{} is the first to bring one-step generation to whole-body control for general humanoid motion tracking.
Training-free guidance optimizes differentiable costs through the generator at inference time~\cite{universalguidance,dps,mpgd}.
Optimizing the input noise of a frozen generator has been used for motion editing (DNO~\cite{dno}) and image generation~\cite{dflow,reno,doodl}; these methods optimize each sample offline, through a multi-step sampler or, in ReNO, a one-step generator, whereas \method{} optimizes the input of a one-step generator inside a 50\,Hz control loop.
In robotics, SWAP~\cite{swap} samples plans above a tracker, and DSRL~\cite{dsrl} steers a diffusion policy by RL in its noise space.
Our steering backpropagates analytic cost gradients through the one-step generator onto its input noise, restricted to selected body-group channels, and requires neither demonstrations nor a learned critic.

\paragraph{Partial references and failure discovery.}
Masked tracking frameworks such as MaskedMimic~\cite{maskedmimic} and AnyBody~\cite{anybody} follow partial pose targets.
In \method{}, masking additionally hands the masked body groups over to a cost while the rest of the body keeps tracking.
To find the motions a tracker fails on, prior work runs the policy in simulation; Extreme-RGMT~\cite{xrgmt}, for example, uses five rollouts per clip over large motion libraries.
We instead use the spread of plans across input noise as a simulation-free estimate of tracking difficulty.

%% file: sec/3_method_new.tex
\section{Method}
\label{sec:method}

\begin{figure*}[t]
  \centering
  \includegraphics[width=\textwidth]{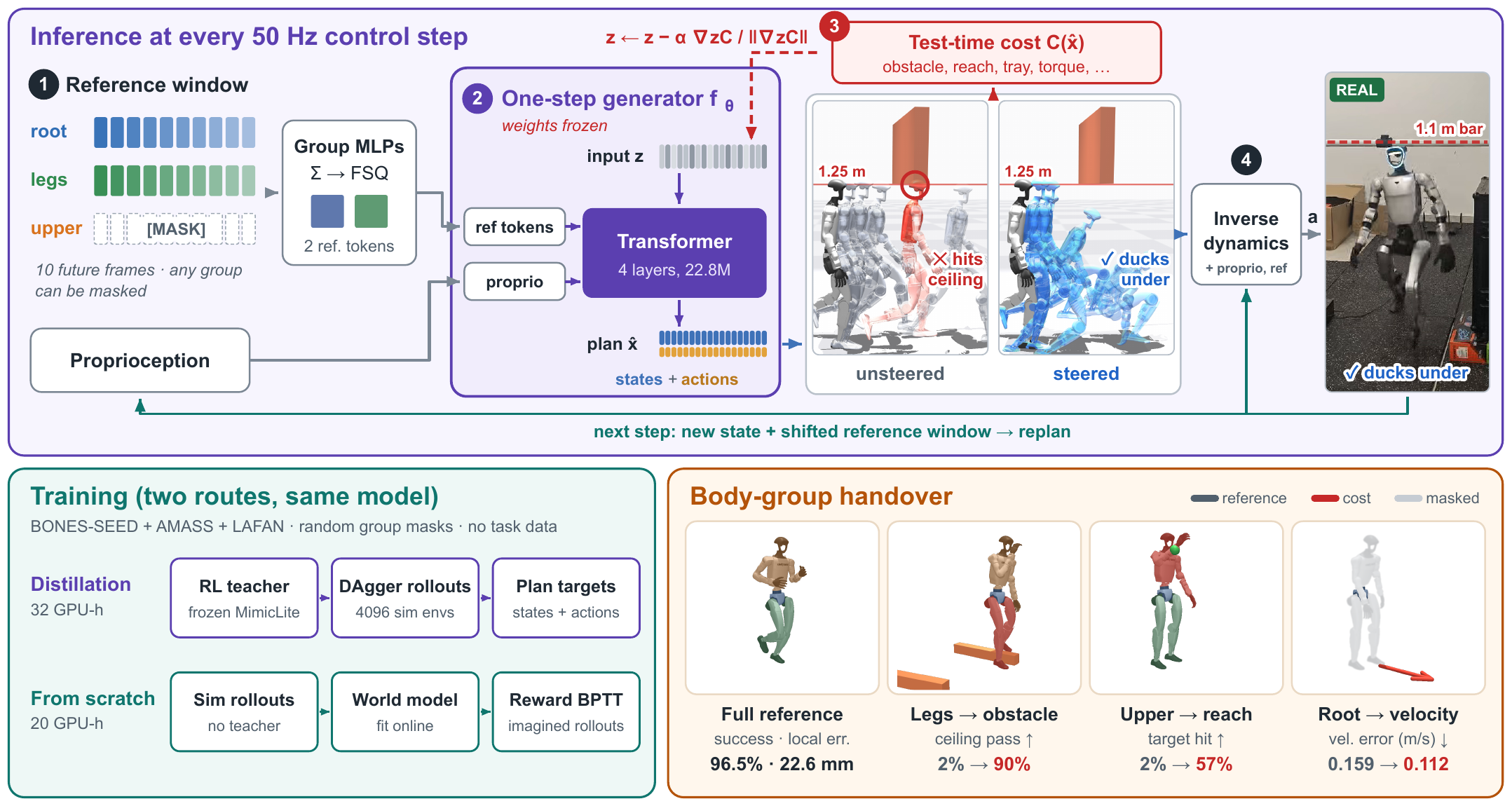}
  \caption{\textbf{Overview of \method{}.}
  \textbf{1} Body-group reference features and proprioception
  condition the controller.
  \textbf{2} A one-step generator predicts a state--action plan.
  \textbf{3} Task costs steer the generator input with weights
  fixed.
  \textbf{4} An inverse-dynamics head computes the current command
  from near-term planned states, current feedback, and the reference.
  The controller replans at the next step.
  Lower panels summarize the two training routes and body-group
  handover (tracking: model trained from scratch, BONES; tasks:
  distilled model, second call steered).
  Plan renderings are simulated; hardware experiments
  evaluate nominal tracking and SDF-based steering.}
  \label{fig:method}
\end{figure*}

\method{} uses a predictive state--action plan as a shared
interface for general tracking, test-time steering, and
feedback-conditioned execution (\cref{fig:method}).
Nominal tracking needs one generator and one head pass;
a new objective steers the generator input by gradient
descent on a cost, with the network weights fixed.
The same predictions also support failure assessment.
Dimensions, frames, constants, cost definitions, the analysis
of the spread, the fine-tuning protocol, and the s.d.\ and run
counts of all results are given in the appendix.

\subsection{Predictive General Tracking}
\label{sec:gen}

\paragraph{Reference conditioning.}
\label{sec:ref}
At control step $k$, the controller receives proprioception
$o_k$ and a future reference window $r_k$ of root poses and
joint angles.
We split each frame into root, legs, and upper-body groups.
A visibility mask $m_{\mathrm{ref}}\in\{0,1\}^{3}$ keeps visible
groups (value 1) as tracking targets and replaces hidden groups
with learned \texttt{[MASK]} embeddings.
Group MLPs and FSQ~\cite{fsq} produce reference tokens, which join
the proprioceptive token to form the \emph{condition}
$c_k=\operatorname{Enc}_{\theta}(o_k,r_k;m_{\mathrm{ref}})$, where
$\operatorname{Enc}_{\theta}$ is the encoder and $\theta$ also
includes the generator's parameters below.
Random masking during training teaches completion
under partial references.

\paragraph{One-step plan generation.}
A plan $X\in\mathbb R^{H\times d}$ has $H$ tokens that sample the
near future densely and later times sparsely.
Each token has $d$ entries: a post-action state (root pose and
velocity, joint positions and velocities, relative to the current
robot pose) and the joint-target action that leads to it.
Given the condition $c_k$ above and an input
$z_k\in\mathbb R^{H\times d}$ with independent standard Gaussian
entries, the transformer generator predicts
\begin{equation}
  \hat X_k=f_\theta(z_k,c_k),
  \label{eq:os-generation}
\end{equation}
where $\theta$ collects the parameters of the generator and the
reference encoder.
One forward pass gives the whole plan, which allows frequent
replanning and a direct differentiable map from generator input to
plan.

\paragraph{Feedback execution.}
The robot does not execute the planned actions.
An inverse-dynamics head with parameters $\phi$ computes the
command as
\begin{equation}
  a_k=g_\phi\bigl(o_k,\hat s_{1:4},r_k;m_{\mathrm{act}}\bigr),
  \label{eq:os-execution}
\end{equation}
where $\hat s_{1:4}$ are the first four planned states of $\hat X_k$
and $r_k$ is the reference window with the groups hidden by a
second mask $m_{\mathrm{act}}\in\{0,1\}^{3}$ zeroed.
The head recomputes the command from the planned states, current
feedback, and the visible reference, so a change to the planned
states changes what the robot does; since it also sees the
reference, it is a plan-conditioned tracker rather than a pure
inverse-dynamics model, and $m_{\mathrm{act}}$ hides the groups it
should not follow.
Only $a_k$ is applied before replanning from updated feedback
and a shifted reference window.
In nominal tracking all groups are visible to both the generator
and the head; the masks are set separately otherwise
(\cref{sec:relax}).

\begin{figure*}[t]
  \centering
  \includegraphics[width=\textwidth]{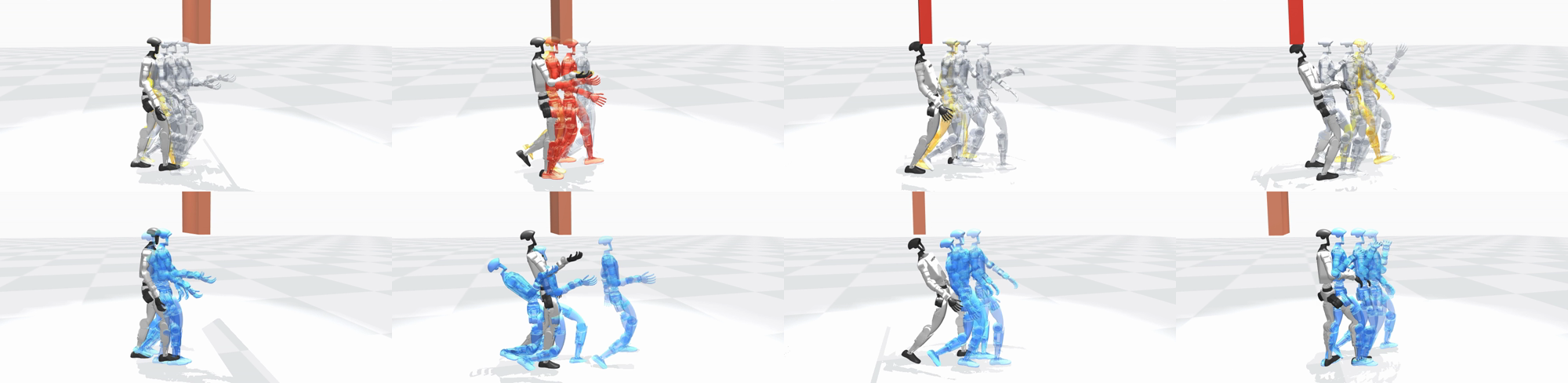}
  \caption{\textbf{Steering moves the plan.} Four time steps (left to right) of one rollout under a 1.25\,m bar. Top: unsteered; grey ghosts show the 0.96\,s plan, red marks planned poses that hit the bar, and gold is the reference. Bottom: the same reference with test-time steering; the plan (blue) ducks under the bar and the robot executes it. Weights, reference and executor are unchanged.}
  \label{fig:ghost_main}
\end{figure*}

\subsection{Test-Time Steering}
\label{sec:steer_method}

\paragraph{Cost-guided input refinement.}
A differentiable task cost $C(X)$ scores a plan, for example by
how deeply it penetrates an obstacle or how far a hand is from a
target.
Since $\hat X=f_\theta(z,c_k)$, the cost can be lowered by changing
the generator input $z$ with the weights fixed (\cref{fig:ghost_main}).
One update moves $z$ by a step of length $\alpha$ against the cost
gradient:
\begin{equation}
  z \;\leftarrow\; z-\alpha\,\frac{G}{\|G\|},
  \qquad
  G=M_{\mathrm{guide}}\odot\nabla_{z}\,C\bigl(f_\theta(z,c_k)\bigr).
  \label{eq:os-guidance-update}
\end{equation}
Here $\odot$ is the elementwise product, $\|\cdot\|$ the Frobenius
norm over the whole tensor, and $M_{\mathrm{guide}}\in\{0,1\}^{H\times d}$
selects the input entries the update may change, for example those
of the legs; since the transformer mixes entries, this restricts
the update, not the plan.
The step length is $\alpha=s\cdot t\cdot\gamma\cdot\sqrt{Hd}$, with guidance
strength $s$, noise level $t$ of the input ($t{=}1$ for pure noise,
$t{=}t_2$ in the second call below), and a factor
$\gamma=\min(1,w/\rho)$ that shortens the step as the plan nears
satisfying the task.
Here $w$ is the largest violation of the task constraint in the
current plan (for example, the deepest penetration) and $\rho$ a
small tolerance, so the push fades once it is met.

\paragraph{Two generator calls.}
Nominal tracking needs one generator call; the controller deployed on the robot keeps the two calls used for steering.
When steering, we call it twice.
The first call turns pure noise into a draft plan $\hat X^{(1)}$.
The second call starts from this draft, mixed with fresh
noise at a level $t_2$ that the generator is trained to
repair~\cite{priming}, as in SDEdit~\cite{sdedit}, and returns the final plan
$\hat X^{(2)}$.
Each call computes $\hat X^{(n)}=f_\theta(z^{(n)},c_k)$ from the input
\begin{align}
  z^{(1)}&=\epsilon\sim\mathcal N(0,I),\\
  z^{(2)}&=(1-t_2)\,\hat X^{(1)}+t_2\,\epsilon',
\end{align}
where $\epsilon'$ is a fresh noise draw.
Before the second call, $J$ updates \eqref{eq:os-guidance-update} are
applied to its input $z^{(2)}$ (steering the first call as well is a
variant), and $\hat X_k=\hat X^{(2)}$ goes to
the inverse-dynamics head.
Steering acts mainly through the second call: the loss-optimal
generator maps every pure-noise input to nearly the same plan
(\cref{sec:difficulty}), so its output responds to the input
mostly when the input carries a draft (1.25\,m bar: 0.02
clean if only the first call is steered, 0.90 if only the second).

\paragraph{Costs.}
A cost only needs to be differentiable with respect to the plan.
Torque costs use the torques implied by the plan's predicted
actions, not the command recomputed by the head.
Costs are given at test time and never seen in training.

\paragraph{Selective reference relaxation.}
\label{sec:relax}
When an objective conflicts with a reference group, hiding that
group removes its tracking target while the others stay as
conditions.
To hand a group to the cost, it must be hidden from the head
($m_{\mathrm{act}}$): a head that still sees the reference follows
it and works against the cost.
Hiding it from the generator too ($m_{\mathrm{ref}}$) is optional.
The guidance mask $M_{\mathrm{guide}}$ is a different object: it
selects the input entries the update may change.
For example, hiding the upper-body reference lets a reaching cost
steer that group while root and legs keep tracking.
Visible groups remain tracking targets, not guarantees: the steered
motion can deviate from them.

\subsection{Training}
\label{sec:train}

Both training routes use BONES-SEED~\cite{bonesseed}, AMASS~\cite{amass},
and LAFAN~\cite{lafan}, without deployment-task demonstrations.
For a clean plan $X^0$, the generator is trained with
\begin{equation}
  \begin{aligned}
    \mathcal L_{\mathrm{gen}}
    &=
    \mathbb E\!\left[
      \|f_\theta(X^{(t)},c)-X^0\|_F^2
    \right],\\
    X^{(t)}&=(1-t)X^0+t\epsilon,
  \end{aligned}
  \label{eq:os-reconstruction}
\end{equation}
where $c$ is the condition $c_k$ of \eqref{eq:os-generation},
$\epsilon$ has independent standard Gaussian entries, and
$t=\operatorname{sigmoid}(u)$ with $u\sim\mathcal N(\mu_u,\sigma_u^{2})$.
As in the one-step recipe of~\cite{priming}, $t$ is not a network
input; in high dimension the noise level is evident from the input~\cite{sun2025noise}.

\paragraph{RL-teacher distillation.}
DAgger~\cite{dagger} gradually transfers control from a frozen
MimicLite teacher~\cite{mimiclite} to the student, and the teacher
labels every visited state.
The plan targets $X^0$ are the future states the robot actually
visited with the teacher's actions.
The head is trained separately by regression onto the teacher's
action, from a perturbed one-step generator sample under random
reference masks; its loss does not update the generator.

\paragraph{Model-based RL from scratch.}
\label{sec:scratch}
A second route learns a world model from simulated transitions
and optimizes tracking and survival objectives through imagined
rollouts with a terminal value critic~\cite{supermimic}, for a
separate instance of the same architecture.
The RL gradient reaches the head but stops at the planned states.
These stay meaningful because the generator is also trained with
$\mathcal L_{\mathrm{gen}}$ on plans actually realized in the replay
buffer (hindsight), and the head with an inverse-dynamics loss on
the true next states.

\subsection{Failure Assessment from Prediction Variation}
\label{sec:difficulty}

\begin{figure}[t]
  \centering
  \includegraphics[width=\columnwidth]{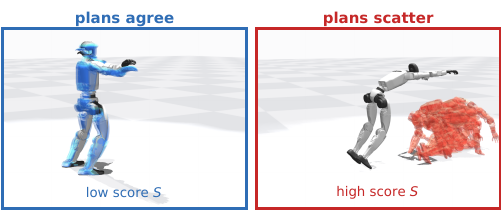}
  \caption{\textbf{Failure assessment from noise spread.}
  The $K$ plans from $K$ noise inputs (colored ghosts) for the same
  reference and state, no simulation.
  Left: the plans agree and the score $S$ is low.
  Right: the plans scatter and $S$ is high.}
  \label{fig:fail}
\end{figure}

The input noise carries no information about the future and, in high
dimension, is recognized as pure noise, so the loss-optimal generator
maps every pure-noise input to nearly the same plan. Variation across
plans from different noise inputs therefore measures how far the trained generator
is from that optimum on the motion (\cref{fig:fail}).
For each clip, we replay the reference as the robot's state, with
no policy rollout.
At each evaluated step $k$, we draw $K$ independent noise inputs
$\epsilon^{1},\dots,\epsilon^{K}$ with proprioception and reference
fixed, and generate $\hat X_k^{(n)}=f_\theta(\epsilon^{n},c_k)$.
Let $\hat q^{(n)}_{k,h,j}$ be the planned position of joint $j\in\mathcal J$
in token $h\in\mathcal H$ of plan $n$, where $\mathcal H$ excludes the
first few tokens.
The spread at step $k$ and the clip score are
\begin{equation}
  \begin{aligned}
    \sigma_k&=\frac{1}{|\mathcal H||\mathcal J|}
    \sum_{h\in\mathcal H}\sum_{j\in\mathcal J}
    \operatorname{std}_{n}\bigl(\hat q^{(n)}_{k,h,j}\bigr),\\
    S&=\operatorname{Quantile}_{0.95}\bigl(\{\sigma_k\}_k\bigr),
  \end{aligned}
  \label{eq:os-spread}
\end{equation}
where $\operatorname{std}_n$ is the standard deviation across the $K$
plans.
High-scoring clips guide targeted fine-tuning with library replay
and a consolidation loss.
The score is an empirical proxy for tracking difficulty, not a
calibrated failure probability; a low score can coexist with a
biased plan.

%% file: sec/4_exp.tex
\section{Experiments}
\label{sec:exp}

\begin{table}[t]
\centering
\small
\setlength{\tabcolsep}{3pt}
\resizebox{\linewidth}{!}{%
\begin{tabular}{@{}lcccc@{}}
\toprule
Method & Succ.\,(\%)$\uparrow$ & $E_l$\,(mm)$\downarrow$ & $E_g$ med.\,(mm)$\downarrow$ & Steer. \\
\midrule
SONIC~\cite{sonic} (release) & \textbf{98.7} / \underline{95.5} & 31.9 / 26.8 & \textbf{99.2} / \underline{56.2} & no \\
SONIC~\cite{sonic} (v1.1) & \underline{98.0} / 95.1 & 30.5 / 27.4 & 126.2 / 67.8 & no \\
HoloMotion~\cite{holomotion} & 95.3 / 94.3 & 43.5 / 38.7 & 119.1 / 96.1 & no \\
HEFT~\cite{heft} (pmg) & 92.7 / 95.3 & \underline{29.4} / \underline{23.3} & 127.0 / 67.1 & no \\
Humanoid-GPT~\cite{humanoidgpt} & 95.3 / 91.0 & 44.5 / 31.8 & 254.3 / 106.1 & no \\
MimicLite Huge~\cite{mimiclite} (teacher) & 94.0 / 95.3 & 29.9 / 27.0 & 138.6 / 68.0 & no \\
\midrule
\method{} (distilled) & 95.3 / 94.9 & 31.4 / 27.3 & 135.5 / 70.0 & \textbf{yes} \\
\method{} (RL from scratch) & 96.7 / \textbf{96.5} & \textbf{26.5} / \textbf{22.6} & \underline{107.6} / \textbf{55.0} & \textbf{yes} \\
\bottomrule
\end{tabular}}
\caption{\textbf{General tracking} in a standardized MuJoCo harness. Each cell is AMASS / BONES (150 and 512 held-out windows). Best in bold, second-best underlined, per dataset. Steer.: steerable at test time. On BONES, 18 windows need stairs, ladders or boxes absent from the flat-ground harness and fail for every method, capping success at 96.5\%. Held out from the teacher's and our training; per their papers, AMASS is in the training data of HoloMotion and Humanoid-GPT and BONES-SEED in that of HEFT.}
\label{tab:track}
\end{table}

We organize our empirical evaluation to answer four central questions:
\textbf{Q1} Can a one-step generative controller match state-of-the-art general tracking fidelity, and can it deploy reliably on physical hardware (\cref{sec:track})?
\textbf{Q2} Does steering inside the generator outperform editing the reference of a deterministic tracker under the same cost (\cref{sec:steer})?
\textbf{Q3} Can the identical frozen controller adapt without training to diverse task objectives and body-group reference masks (\cref{sec:more})?
\textbf{Q4} Can the spread of plans diagnose tracking failures without simulation, and does fine-tuning on flagged motions repair them (\cref{sec:diffexp})?

\subsection{Q1: General Tracking Fidelity, Dual Training, and Physical Hardware Deployment}
\label{sec:track}
\paragraph{Protocol.}
All trackers run in the same MuJoCo~\cite{mujoco} sim-to-sim harness with their official policies at 50\,Hz, on 150 held-out windows of AMASS~\cite{amass} subjects and all 512 held-out windows of BONES-SEED actors.
Following SONIC~\cite{sonic}, a clip fails if the pelvis height deviates from the reference by more than 0.25\,m; on successful clips we report the root-relative error $E_l$ and the median global error $E_g$.

\paragraph{Results.}
\cref{tab:track} benchmarks \method{} against five competitive general-purpose humanoid trackers (six checkpoints).
Distilled \method{} performs on par with its RL teacher (local error $31.4$ vs.\ $29.9$\,mm on AMASS, $27.3$ vs.\ $27.0$\,mm on BONES); its larger global error than SONIC on AMASS comes from drift on near-static poses, as for the teacher.
Trained from scratch (\cref{sec:scratch}), \method{} has the lowest local tracking error of all six baselines on both benchmarks ($26.5$\,mm on AMASS, $22.6$\,mm on BONES) and the highest success and lowest median global drift on BONES, so the generative formulation costs no local tracking accuracy.
One generator call tracks as well as two (in the harness of \cref{tab:track}, 96.0\% / 94.9\% success with either, $E_l$ 31.9 vs.\ 32.0 / 27.2 vs.\ 27.3\,mm), so the second call is free to serve steering (\cref{sec:steer_method}).
Steering (Q2, Q3) uses the distilled model, so that it is compared with editing the teacher's reference in the same setting.

\begin{figure*}[t]
  \centering
  \setlength{\tabcolsep}{0pt}
  \begin{tabular}{@{}c@{\hspace{3pt}}c@{\hspace{8pt}}c@{}}
  & {\small Passing under a 1.1\,m bar} & {\small Stepping over a 12\,cm hurdle} \\[2pt]
  \rotatebox{90}{\scriptsize\hspace{1pt}Without steering} & \includegraphics[width=0.485\textwidth]{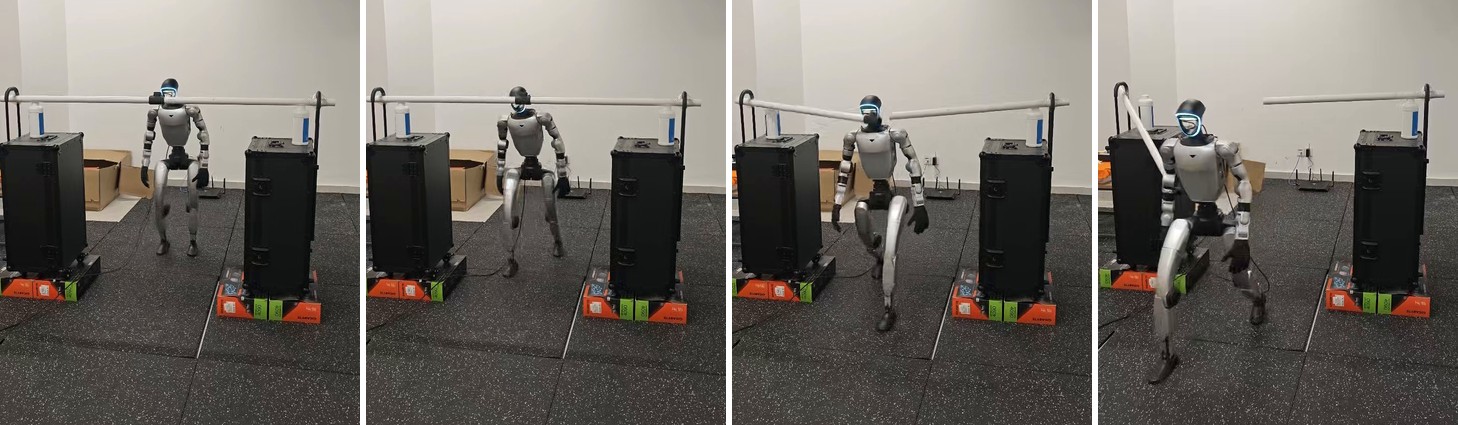} & \includegraphics[width=0.485\textwidth]{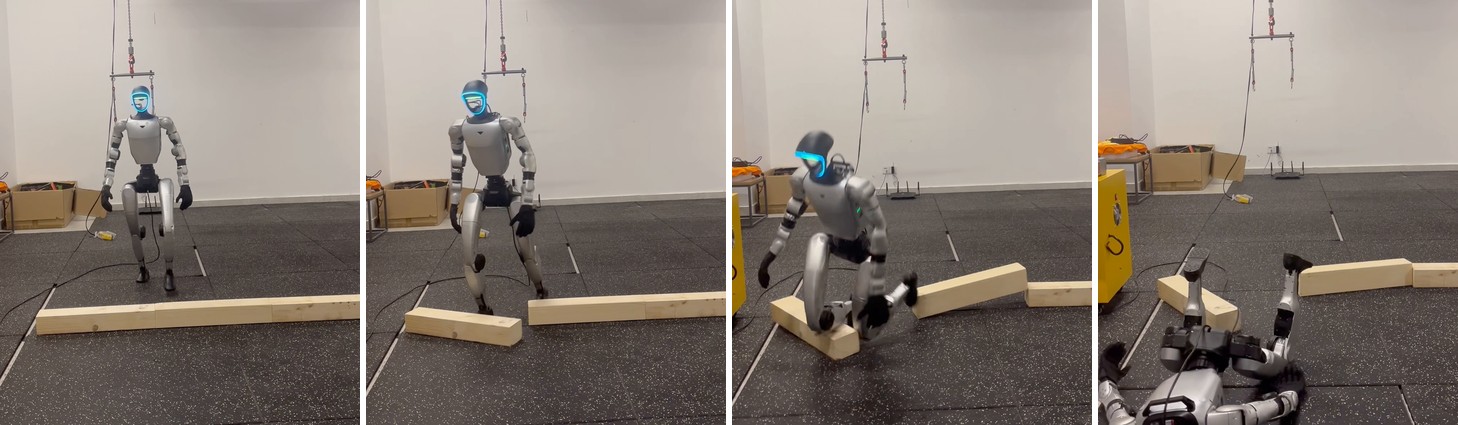} \\[1pt]
  \rotatebox{90}{\scriptsize\hspace{7pt}With steering} & \includegraphics[width=0.485\textwidth]{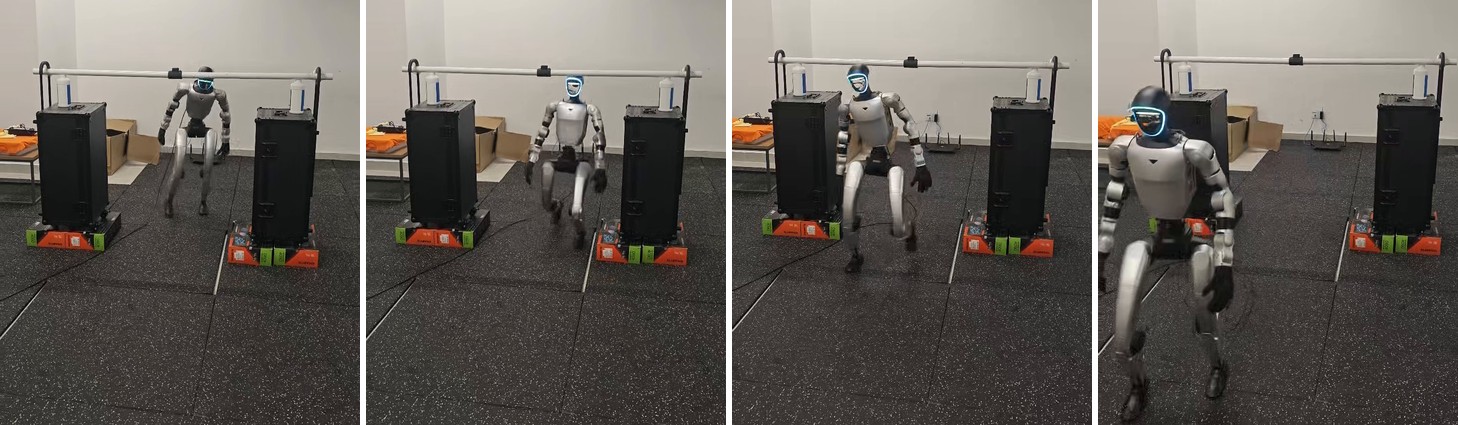} & \includegraphics[width=0.485\textwidth]{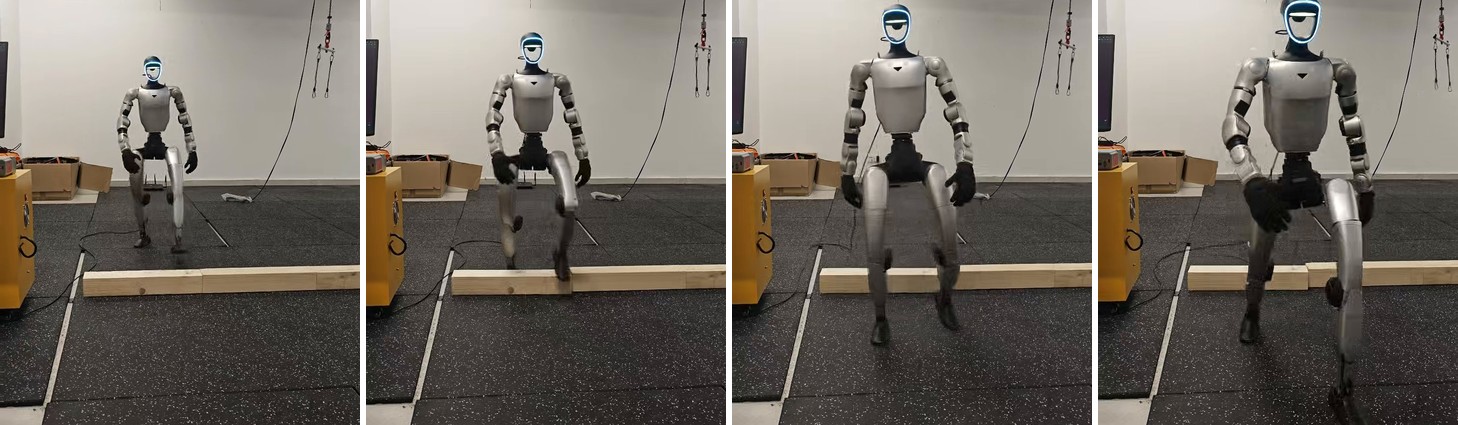} \\
  \end{tabular}
  \caption{\textbf{Steering on a physical Unitree G1} with an SDF cost built from the onboard LiDAR and the same frozen policy throughout. Left: unsteered, the head hits the bar and knocks it off; steered, the robot lowers early, passes under, and walks out to stand. Right: unsteered, the foot catches the hurdle and the robot falls; steered, the robot raises each foot and steps over.}
  \label{fig:real_steer}
\end{figure*}

\paragraph{Latency and real-robot deployment.}
On an RTX 4090 GPU, a nominal tracking step takes 3.8\,ms per generator call (5.3\,ms for the robot's two calls).
On the robot's Jetson Orin NX, timed offline under the same measurement, tracking with two calls takes 4.2\,ms and steering the second call or both calls 14.4 or 18.2\,ms (\supp{tab:latency}), within the 20\,ms period of the 50\,Hz loop.
On a physical Unitree G1, the distilled \method{}, exported to ONNX, completed all 27 trials (nine whole-body motions, three each) without a fall.
With an SDF cost from a fixed LiDAR map built before each run, it passed under a 1.1\,m bar in four of five attempts and stepped over a 12\,cm hurdle without falling in all three (once without touching it), where the unsteered controller failed every attempt (\cref{fig:real_steer}); in these trials the steered policy ran on a tethered RTX 4090 computer.

\begin{table}[t]
\centering
\small
\setlength{\tabcolsep}{3pt}
\resizebox{\linewidth}{!}{%
\begin{tabular}{@{}lccccccc@{}}
\toprule
 & \multicolumn{2}{c}{Hurdles$\uparrow$} & \multicolumn{3}{c}{Low bar: clean pass$\uparrow$} & Reach & Latency \\
\cmidrule(lr){2-3}\cmidrule(lr){4-6}
Method & feasible & clean & 1.30\,m & 1.25\,m & 1.20\,m & hit$\uparrow$ & (ms)$\downarrow$ \\
\midrule
Teacher, original ref. & 0.48 & 0.01 & 0.36 & 0.02 & 0.00 & 0.01 & 2.5 \\
Teacher, offline edit (J) & 0.58 & 0.01 & 0.72 & 0.06 & 0.00 & 0.38 & 2.5 \\
Teacher, offline edit (J+R) & 0.47 & 0.00 & 0.55 & 0.03 & 0.00 & 0.31 & 2.5 \\
Teacher, closed-loop edit (J) & 0.57 & 0.02 & 0.87 & 0.11 & 0.02 & \underline{0.60} & 330.5 \\
Teacher, closed-loop edit (J+R) & 0.45 & 0.00 & 0.67 & 0.04 & 0.02 & \underline{0.60} & 277 \\
Teacher, tracks our plan (J) & 0.53 & 0.10 & 0.86 & 0.74 & 0.51 & 0.37 & 10.6--13.7 \\
\midrule
\method{}, unsteered & 0.40 & 0.00 & 0.23 & 0.02 & 0.00 & 0.02 & 5.3 \\
\method{}, 2nd call, 2 steps & \underline{0.68} & 0.11 & \underline{0.95} & \textbf{0.90} & \textbf{0.56} & 0.57 & 10.6--13.7 \\
\method{}, both calls, 4 steps & 0.66 & \underline{0.12} & \underline{0.95} & \underline{0.84} & \underline{0.52} & 0.54 & 17.2--20.5 \\
\method{}, 6 steps + line search & \textbf{0.74} & \textbf{0.21} & \textbf{0.98} & 0.83 & \textbf{0.56} & \textbf{0.62} & 100--112 \\
\bottomrule
\end{tabular}}
\caption{\textbf{Steering inside the generator versus editing the reference} (79 walking windows). The teacher tracks a reference edited under the same cost, offline or in closed loop; J: task-group joints, R: also root height and pitch. Tracks our plan: a planner followed by a separate tracker, the teacher tracking the plan of our 2nd-call row (same steering computation). 6 steps + line search: in each call, a larger optimization budget. Hurdles: feasible passes may touch the hurdle with a foot, clean passes do not. Latency per control step on one RTX 4090. Bold: best, underlined: second best per column. Our 2nd-call and both-calls rows stay within the 20\,ms period of the 50\,Hz loop on the robot's Orin NX (95th percentile).}
\label{tab:steer}
\end{table}

\subsection{Q2: Generative In-Controller Steering vs.\ Kinematic Reference Editing}
\label{sec:steer}
\paragraph{Tasks and baselines.}
We steer 79 walking motions of unseen actors past 10\,cm hurdles (feasible pass), under a low bar at 1.30, 1.25 or 1.20\,m (clean pass) and a wrist target (hit), three runs each.
A deterministic tracker can instead edit its reference under the same cost; the teacher tracks a reference edited offline or re-optimized in closed loop.

\paragraph{Results.}
On obstacles, steering inside the generator beats every reference editor in \cref{tab:steer}, including editors re-optimized from the robot's state in closed loop and editors free to move the root: by a wide margin under the low bar, and by a smaller one on hurdles. A planner followed by a separate tracker also falls short: the teacher tracking our own steered plan passes the 1.25\,m bar in 0.74 of the trials instead of 0.90, so the gain also relies on the head trained with the generator.
In the real-time setting used on the robot (second call guided), \method{} passes 0.68 of the hurdles against at most 0.58 for editing, and under the 1.25\,m bar 0.90 against at most 0.11.
The cause is execution, not planning: the edited references clear the bar, but where the edit lowers the pelvis by 6.2\,cm the teacher realizes 1.0\,cm, as a tracker trained on natural motions does not follow such edits; steered \method{} lowers its pelvis by 5.9--6.6\,cm, since its executor is trained to follow its own plan (\cref{fig:ghost_main}).
In reaching, real-time steering (0.57 hits) far exceeds every real-time editor (at most 0.38) and nearly matches closed-loop editing (0.60) at under a twentieth of its computation; 12 smaller steps reach 0.75.
Controls attribute the gain to the cost gradient on the generator input (\supp{tab:mech}): a random update of equal norm and best-of-eight sampling yield little improvement; supplying \method{} with the edited reference yields gains comparable to those of the teacher; and optimizing the generated plan directly improves reaching but degrades obstacle passage and loses the tray's ball in nearly every window.
Our steering covers generators from one step to many: a multi-step flow version of our model is steered the same way with comparable task results, while the one-step generator makes nominal tracking cheaper and steering smoother (\supp{tab:flow}).
The settings, chosen on these windows, carry over unchanged to 79 held-out walking windows and to held-out dance.

\subsection{Q3: Training-Free Multi-Objective Adaptation and Body-Group Handover}
\label{sec:more}

\begin{figure}[!t]
  \centering
  \includegraphics[width=0.85\linewidth]{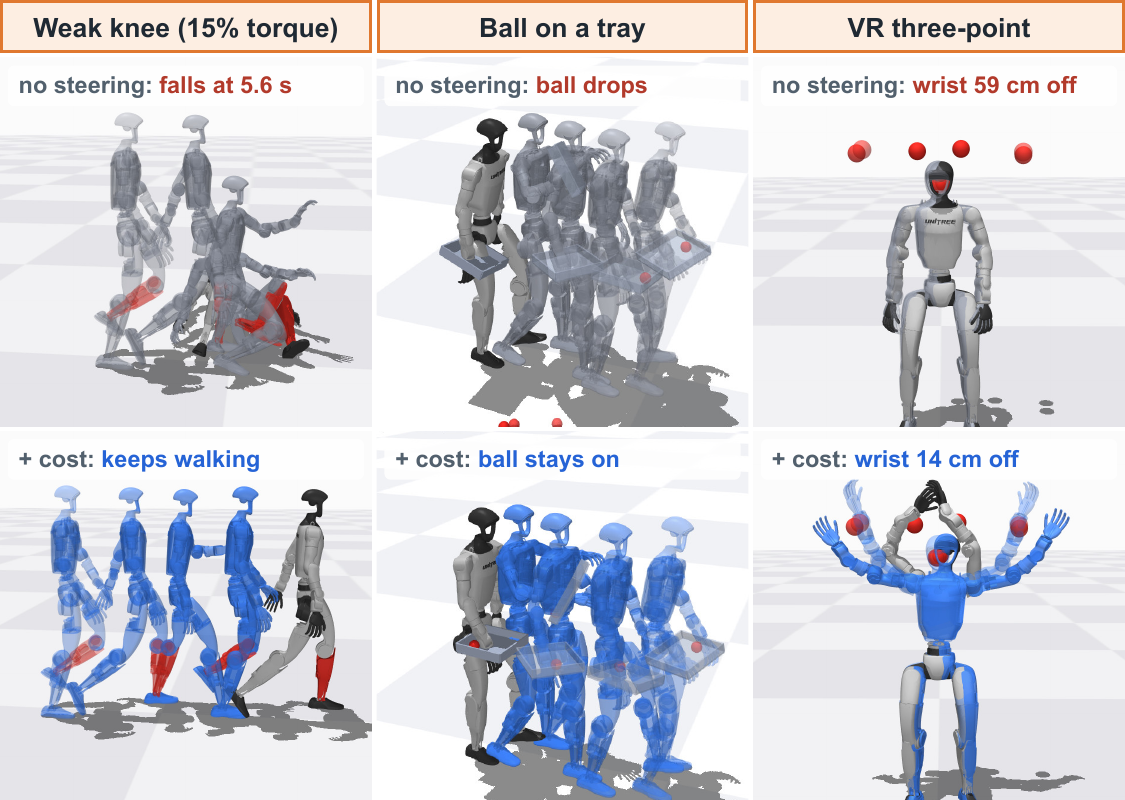}
  \caption{\textbf{The same frozen policy without (top) and with (bottom) a test-time cost} (simulation; one clip each, same reference; earlier poses in grey without and blue with steering). Over the test sets, steering raises knee-fault survival from 0.71 to 0.79, halves ball loss (0.16 to 0.08) and lowers the VR wrist error from 24.3 to 16.4\,cm.}
  \label{fig:q3}
\end{figure}

\paragraph{Multi-objective training-free steering.}
Without fine-tuning or demonstrations, the same frozen \method{} adapts to new objectives through a cost (\cref{fig:q3}).
\emph{Actuator faults}: with a knee or hip motor limited to 15\% of its torque and only the cost aware of the defect, penalizing torques above the degraded limit raises survival from $0.71{\pm}0.01$ to $0.79{\pm}0.01$ (knee) and from $0.79{\pm}0.02$ to $0.84{\pm}0.01$ (hip).
\emph{Payload}: holding a tray with a free rolling ball, a cost on the hand's tilt halves ball loss (0.16 to 0.08).
\emph{Velocity commands}: with the leg and upper-body references masked, a velocity cost turns \method{} into an omnidirectional velocity policy (error 0.159 to 0.112\,m/s).

\paragraph{Body-group handover and sparse tracking.}
Masking a reference group hands its degrees of freedom to the cost, which matters when the reference opposes the cost (\supp{tab:mask}).
For a velocity command the reference is a standing pose, so a tracker that still sees it shuffles: hiding the legs and upper body from the generator alone barely changes the error, whereas hiding them from the executor as well lowers it from 0.218 to 0.159\,m/s.
Where the reference does not oppose the cost, masking is harmless: with legs masked, \method{} clears hurdles and low bars at the same rates as full tracking, nominal survival is unchanged without costs, and the visible groups keep following the reference.

\subsection{Q4: Simulation-Free Failure Discovery and Targeted Fine-Tuning}
\label{sec:diffexp}

\begin{figure}[t]
  \centering
  \vspace{-4.5mm}
  \includegraphics[width=\linewidth]{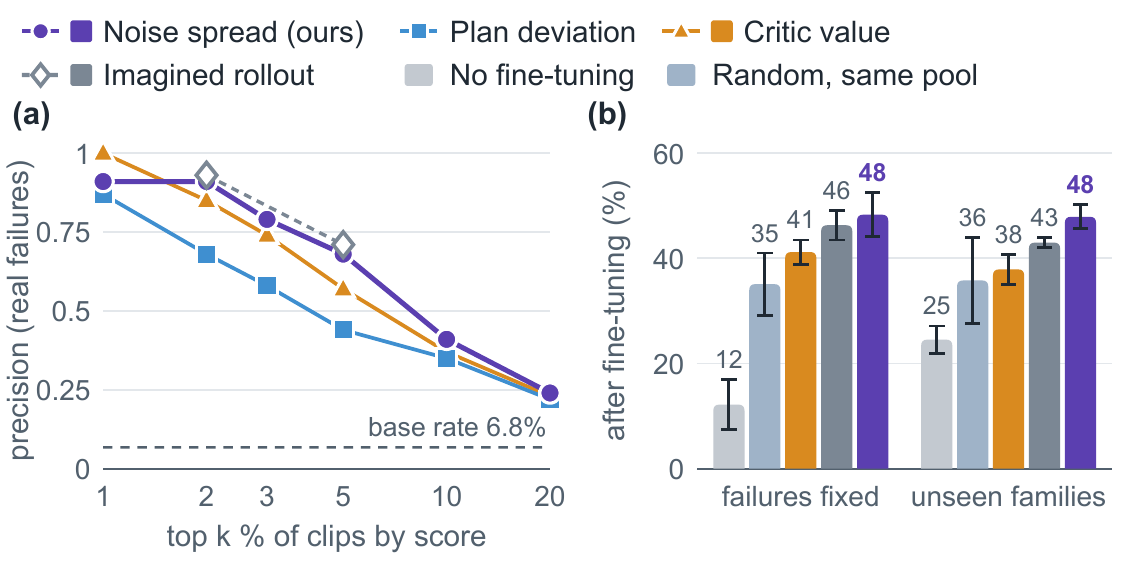}
  \caption{\textbf{Simulation-free failure discovery and repair.} (a) Fraction of real failures among the top $k\%$ of 7{,}884 clips ranked by each score; imagined rollouts are reported only at 2\% and 5\% and need a world model. (b) Fine-tuning on 329 clips selected by each score or drawn at random from the same candidate pool, same recipe: share of base-model failures fixed, and success on 110 held-out failures from unseen skill families; bars show the mean over three training seeds and whiskers ${\pm}1$ s.d.}
  \label{fig:selfassess}
  \vspace{-0.4mm}
\end{figure}

Across the 7{,}884 training-library sequences, the nominal tracker fails on 537 clips (6.8\%); held-out skill families test generalization below.
Among the top 2\% of clips ranked by the spread of plans across noise draws, 91\% are real failures (90\% when whole clips are scored), near the 93\% of imagined world-model rollouts, with no model beyond the generator (\cref{fig:selfassess}a); on the distilled model, which has no world model, it reaches 86\% (base rate 8.4\%). Over the whole ranking the spread trails an RL critic and imagined rollouts (AUC 0.83 against 0.86 and 0.91), which need a learned value function or world model; at the top of the ranking, which is what selection uses, it comes close to imagined rollouts.
Fine-tuning on the 329 clips it selects (\cref{fig:selfassess}b; three training seeds) fixes $48{\pm}4\%$ of the base model's 501 failures on a rerun of the library (7{,}788 clips) (rerunning alone: $12{\pm}5\%$; outside its own 329 clips: $44{\pm}6\%$), above the $35{\pm}6\%$ of 329 random clips from the same candidate pool ($27{\pm}2\%$ from the whole library) and the $41{\pm}2\%$ of clips ranked by an RL critic under the same recipe, and on par with 329 failures found by one simulated rollout per clip ($51{\pm}4\%$), while regressions stay at $1.8{\pm}0.2\%$ and the local error on 40 hard clips the base model tracks rises only slightly ($35.3$ vs.\ $33.6$\,mm).
The repair generalizes beyond the selected clips: success on 110 held-out failures from unseen skill families rises to $48{\pm}2\%$ (random: $35{\pm}4\%$; critic: $38{\pm}3\%$), above clips selected by imagined rollouts ($43{\pm}1\%$), which need policy rollouts and a world model.
Scoring needs no simulator and, on whole clips, is 11--28$\times$ cheaper than the five simulated rollouts per clip that Extreme-RGMT~\cite{xrgmt} uses for a reliable label, and 2.4$\times$ cheaper than one rollout per clip (\supp{sec:app_self}).

%% file: sec/5_conclusion.tex
\section{Conclusion and Future Work}
\label{sec:concl}

We presented \method{}, to our knowledge the first one-step diffusion whole-body controller for general humanoid tracking. Trained from scratch, it has the lowest local tracking error among the evaluated general trackers, and its distilled version runs on a Unitree G1; with frozen weights, costs on its plan and body-group masks adapt it to new objectives while the unmasked reference is still followed; and the spread of its plans finds tracking failures without simulation to guide fine-tuning.
\paragraph{Future work.}
\emph{Vision}: depth images or egocentric views can join the condition as tokens next to the reference, as the height scan does in our terrain variant, so that the plan itself anticipates obstacles and uneven ground; a cost could then also be read from the camera, \eg a distance field predicted from images instead of a pre-built map.
\emph{Force control}: pushing, carrying or opening doors require sustained contact forces, which a kinematic plan does not represent; adding wrench or tactile tokens to the condition and contact forces to the plan would let a cost prescribe a force instead of a pose.
In both cases the generator, steering and failure assessment carry over unchanged.

%% file: sec/6_appendix.tex
\section{Implementation details}
\label{sec:app_impl}
\cref{tab:impl} lists the architecture, training, steering and evaluation settings that the main text refers to.

\paragraph{Steering costs.}
\label{sec:app_costs}
The costs used in the experiments are:
(i)~\emph{obstacle avoidance}, with collision capsules on the G1 evaluated by differentiable forward kinematics against the signed-distance field (SDF) of oriented boxes;
(ii)~\emph{reaching}, a hinge loss on the distance between an end effector and a target;
(iii)~\emph{pelvis-relative drawing}, the distance of a hand to prescribed 3D curves in the pelvis frame;
(iv)~\emph{weak motors}, a soft penalty on joint torques $k_p(q_{\text{def}}+\kappa a-q)-k_d\dot q$ above the degraded motor limits, with action scale $\kappa$;
(v)~\emph{tray balancing}, a penalty on the tilt of a hand-held tray from vertical;
(vi)~\emph{velocity commands}, a planar root-velocity error with foot-slip regularization. The commands are 64 synthetic clips of 10\,s: 1\,s of standing, then low-pass-filtered segments of 1.5--3\,s (forward at 0.3--1.2\,m/s, backward, sideways at 0.2--0.4\,m/s, turning at 0.5--1.0\,rad/s, walking turns and stops). The reference root is integrated from the command and its joints are a standing pose without a gait. The reported error is the planar velocity error in the body frame from 0.2\,s finite differences, skipping 0.5\,s after each change of command.

\begin{table*}[t]
\centering
\footnotesize
\setlength{\tabcolsep}{3pt}
\begin{minipage}[t]{0.485\textwidth}
\centering
\begin{tabular}[t]{@{}l>{\raggedright\arraybackslash}p{0.52\linewidth}@{}}
\toprule
\multicolumn{2}{@{}l}{\textbf{Reference interface}} \\
Window & 10 future frames, 0.1\,s apart \\
Frame features & 38 (root 9, legs 12, upper body 17) \\
Reference tokens & 2 FSQ tokens of 32 dims \\
Mask prob.\ (group / window) & 0.3 / 0.1 \\
\midrule
\multicolumn{2}{@{}l}{\textbf{Generator and action head}} \\
Transformer & 4 layers, width 512, 8 heads \\
Parameters & 22.8M \\
Prediction tokens & $H{=}24$: 16 dense + 8 every 4 ticks (0.96\,s) \\
Token content & state 73 + action 29 \\
Training noise scale & $t=\operatorname{sigmoid}(u)$, $u\sim\mathcal N(0.8,0.8^{2})$ \\
Action-head input & proprioception + $\hat s_{1:4}$ + reference window + group mask $m_{\mathrm{act}}$ \\
\midrule
\multicolumn{2}{@{}l}{\textbf{Data}} \\
Motion library & BONES-SEED, AMASS, LAFAN \\
Size & 78{,}842 sequences, $\approx$170\,h \\
\midrule
\multicolumn{2}{@{}l}{\textbf{Online distillation}} \\
Teacher & MimicLite Huge \\
Parallel environments & 4096 \\
DAgger $\beta$ & 1$\to$0 over 30 iterations \\
Reconstruction loss weight & 0.01 \\
Schedule & 2390 it., 8\,h on 4$\times$RTX 4090 (32 GPU-h) \\
\midrule
\multicolumn{2}{@{}l}{\textbf{Training from scratch}} \\
Parallel environments & 4096 \\
World model & 3-layer MLP \\
Imagined rollout & 32 steps \\
Schedule & 60{,}000 it.\ ($\approx$20 GPU-h) \\
\bottomrule
\end{tabular}
\end{minipage}\hfill
\begin{minipage}[t]{0.485\textwidth}
\centering
\begin{tabular}[t]{@{}l>{\raggedright\arraybackslash}p{0.52\linewidth}@{}}
\toprule
\multicolumn{2}{@{}l}{\textbf{Self-assessment and repair}} \\
Noise draws & $K{=}8$, evaluated at 10\,Hz \\
Score & 95th pct.\ over time of joint-angle std \\
Cost & 0.07--0.09\,s per clip on one GPU \\
Fine-tuning set & top 329 flagged clips after exclusions \\
\midrule
\multicolumn{2}{@{}l}{\textbf{Steering}} \\
Default strength $s$ & hurdles 1, low bar 4, reaching 2 \\
Collision geometry & 124 capsules vs.\ box SDF \\
Latency & RTX 4090 with CUDA graphs \\
\midrule
\multicolumn{2}{@{}l}{\textbf{Feasibility projection}} \\
Horizon & $K{=}12$ actions \\
Optimizer & Adam, 10 iterations \\
World model & 3-layer MLP, 2048 hidden, 8.8M params \\
World-model data & 5.0M transitions of \method{} rollouts \\
\midrule
\multicolumn{2}{@{}l}{\textbf{Evaluation}} \\
Tracking test sets & 150 (AMASS) / 512 (BONES) windows of up to 10\,s, fixed seeds \\
AMASS split & 33 subjects excluded from training \\
BONES split & evaluation actors excluded from training \\
Teacher data & the same exclusions \\
Reach target & 15\,cm forward, 15\,cm up; radius 8\,cm \\
Reach time window & $\pm0.25$\,s mid-sequence \\
Edited groups (baselines) & hurdles: legs; low bar: legs, torso; \\
 & reach: right arm, torso \\
Fault success (\cref{tab:mask}) & mean of 2--3 runs on 150 clips \\
\bottomrule
\end{tabular}
\end{minipage}
\caption{\textbf{Implementation details.}}
\label{tab:impl}
\end{table*}

\paragraph{Method details.}
\label{sec:app_method}
\emph{Constants.}
$H=24$ plan tokens covering 0.96\,s; $d=102$ entries per token (73 state, 29 action); $J=2$ updates per call; second-pass noise level $t_2=0.8$; tolerance $\rho=0.05$ (in the unit of $w$); default strength $s$ per task (\cref{tab:impl}); training noise level $t=\operatorname{sigmoid}(u)$ with $u\sim\mathcal N(\mu_u,\sigma_u^2)$, $\mu_u=\sigma_u=0.8$; $K=8$ noise samples and a 95th-percentile score; the reference window has 10 frames, 0.1\,s apart.
\emph{Reference window.}
The window has 10 frames at offsets of 0, 5, \dots, 45 control ticks (0.1\,s apart).
Each frame has 38 features: root 9 (position in the yaw frame of the current reference frame, and a 6-number orientation relative to the robot), legs 12 and upper body 17 (joint positions).
\emph{Plan.}
The 24 tokens sit at tick offsets 0--15 and then every fourth tick up to 47 (about 0.96\,s).
A token has $d=102$ entries: a state of 73 (root position 3, orientation 6, linear velocity 3, angular velocity 3, joint positions 29, joint velocities 29) and an action of 29.
States are relative to the current robot pose: root $xy$ rotated into the robot's yaw frame, root height absolute, orientation relative to the current root, linear velocity in the yaw frame, angular velocity in the body frame.
Plan and reference features are normalized with their dataset means and standard deviations; positions are in metres, angles in radians, and velocities in m/s and rad/s.
Costs that act on world-frame quantities, such as obstacle distances, transform the plan back with the current pose.
\emph{Head.}
$g_\phi$ is an MLP (hidden widths 1024, 1024, 512, Mish) whose input is proprioception, the first four planned states, the reference window and the three group flags; a hidden group's reference features are set to zero.
It is trained with group masks drawn per group with probability 0.5 and for all groups with probability 0.2, whereas the generator's reference masks use 0.3 and 0.1.

\emph{Steering update.}
The gradient is divided by $\max(\|G\|_F,10^{-6})$, with the norm taken over the whole $H\times d$ tensor after masking, so with a channel mask the active channels share the full step; a zero gradient gives a zero update.
The violation $w$ that sets $\gamma=\min(1,w/\rho)$ is, per cost, the largest value over the plan of: the clearance violation of the whole-body capsules against the obstacle field (obstacle avoidance, m); the distance of the end effector to the target, beyond a 5\,cm dead zone for reaching and without it for pelvis-relative drawing (m); and the tilt angle of the tray (rad, first 16 plan ticks).
For velocity commands, $w$ is the root-mean-square planned velocity error (m/s).
Weak-motor and other resource costs (energy, quiet landing, support) have no constraint to satisfy and always take the full step ($\gamma=1$).
The same $\rho=0.05$ is used in the unit of $w$.
The step length is set by the violation only; regularizing terms in $C$ change the direction but not the length.
The second-pass input is $0.2\,\hat X^{(1)}+0.8\,\epsilon'$, the same corruption as in training, $X^{(t)}=(1-t)X^0+t\epsilon$ at $t=0.8$, with fresh noise $\epsilon'$.

\emph{Steering cost.}
With both calls steered by two steps each, one control step evaluates the generator six times, four of them with a backward pass (four and two with the second call only), and the cost is not re-evaluated after the last step.
The line-search variants (\cref{sec:steps}) halve a step up to $n$ times and add one evaluation per halving.
The settings with a larger optimization budget run 12 steps, or 6 steps with a line search, in each call (\cref{sec:steps}).
Torque costs penalize the torque of the weak joints implied by the planned actions of the first 16 ticks (hinge squared above the reduced limit); a variant evaluates the action that the head would execute now, and is used in one setting only. Only the actions of the first 8 ticks are supervised in training (\cref{sec:app_method}); restricting the cost to them gives the same result (left knee at 15\%, in a separate batch: success $0.79{\pm}0.01$ with 8 ticks and $0.78{\pm}0.01$ with 16, against $0.79{\pm}0.01$ with 16 ticks in \cref{tab:more}; 0.71 unsteered; three runs).

\emph{Consistency refinement.}
The world model is an MLP (three hidden layers of 2048), trained once and offline on closed-loop rollouts of the distilled controller over 6{,}000 training clips (action noise 0, 0.1 and 0.3) without task data; it is separate from the world model that the scratch-RL route fits online.
Starting from the measured state, we optimize offsets to the planned actions of the first $K$ ticks ($K=8$ or 12) with Adam (10 steps, learning rate 0.1) to match the steered joint positions, the root height and the root tilt of the plan.
Errors are scaled by 0.05\,rad, 0.02\,m and 0.03, and the offsets carry an $\ell_2$ penalty of 0.01.
The rollout replaces the first $K$ planned states before the head.

\emph{Training.}
In distillation, the executed action comes from the teacher with probability $\beta$, annealed from 1 to 0, and otherwise from the student; half of the environments add action noise of 0.1 (DART).
Besides $\mathcal L_{\mathrm{gen}}$, the generator is trained with a small reference reconstruction loss (straight-through gradients through the quantizer) and an action-difference smoothness term (weight 0.1, first eight plan ticks).
The head's input states are a one-step generator sample plus Gaussian noise of 0.02 (normalized units).
In scratch RL, the head's inverse-dynamics loss uses the true next states with noise 0.1 (normalized units) and a weight of 1; the generator's hindsight loss has weight 1, and a small reference reconstruction loss (weight 0.01) is added.
The generator's action channels are supervised only on the first eight plan ticks, since only the head produces the executed action.

\emph{Failure score.}
In symbols, let $\hat q^{(n)}_{k,h,j}$ be the planned position of joint $j$ in plan token $h$ at step $k$ for noise sample $n=1,\dots,8$.
The spread at step $k$ is $\sigma_k=\operatorname{mean}_{h,j}\operatorname{std}_n\,\hat q^{(n)}_{k,h,j}$, and a clip's score is the 95th percentile of $\sigma_k$ over $k$; $h$ runs over the plan tokens from the fifth onward.
The reference clip supplies the state fed to the controller (teacher forcing), and the score is computed every fifth tick over the first 500 ticks.
The standard deviation across the $K=8$ plans is the population one (divided by 8), in radians, over joint positions only, for plan tokens from the fifth onward.

\section{Evaluation protocol}
\label{sec:app_eval}
\paragraph{Tracking protocol.}
We evaluate all trackers within a unified simulation benchmark: MuJoCo~\cite{mujoco} sim-to-sim derived from the MimicLite deployment harness, executing each method's official ONNX policy and control configuration at 50\,Hz. Following SONIC~\cite{sonic}, an episode is classified as failed if the vertical pelvis position deviates from the reference by more than 0.25\,m at any frame. Tracking metrics are aggregated over successful clips: root-relative body joint position error $E_l$ (mm) and global root-translated body position error $E_g$ (mm). Owing to the heavy-tailed distribution of global drift, we report the median of $E_g$. Evaluation spans two test sets: 150 windows of held-out subjects from AMASS~\cite{amass} and all 512 windows of unseen actors from BONES-SEED (\cref{tab:impl}). On BONES, 18 windows (50\,cm boxes, ladders, stairs, a 1\,m jump) require scene geometry absent from the flat-ground harness and fail for every tracker; the RL-from-scratch \method{} fails only these.

\paragraph{Steering protocol.}
Steering capabilities are evaluated in the training simulator (mjlab, on MuJoCo Warp), not in the MuJoCo harness used for \cref{tab:track}, on 79 walking trajectories from unseen actors across three random trials per condition: (i)~\emph{Hurdles}: three 10\,cm crossbars placed across the locomotion path at 1, 2, and 3\,m intervals. Crossings are categorized as \emph{feasible passes} if hurdles are cleared, stepped upon, or lightly contacted without falling, and as \emph{severe failures} if the torso impacts the barrier or the robot collapses. (ii)~\emph{Low bar}: a horizontal beam (10\,cm deep along the path, 80\,cm wide, 50\,cm tall) 1.5\,m along the path, with its underside at 1.30, 1.25, or 1.20\,m, with \emph{clean passes} denoting collision-free traversal under the beam: a crossing beside it (root lateral offset beyond its half-width of 0.4\,m) is not clean. Such detours are 1.5--7.6\% of the encounters in every row of \cref{tab:steer}, steered, unsteered and edited alike, and the mean lateral offset at the beam is 8--18\,cm. (iii)~\emph{Spatial reaching}: the right wrist must reach a target offset forward and upward from its nominal reference position within a mid-sequence time window. Target geometry and default guidance strengths are listed in \cref{tab:impl}.

\paragraph{Baselines and controls.}
The established alternative for deterministic controllers is to optimize kinematic reference modifications under identical cost objectives prior to tracking. We optimize joint-level trajectory edits of the task-relevant body groups, preserving root kinematics (\cref{tab:impl}), under identical signed-distance or reaching costs alongside temporal smoothness regularizers. The modified trajectories are subsequently executed by the frozen RL teacher (MimicLite Huge). We benchmark against both offline kinematic editing and closed-loop kinematic editing, where the remaining 1.2\,s horizon of the reference trajectory is re-optimized from the robot's measured root state at every 50\,Hz control step for hurdles and reaching and every 5 steps for the low bar (every step gives 0.10 instead of 0.11 at 1.25\,m).
Because low-bar steering can lower the root while these editors keep it, we add for the low bar an editor that also changes the root height and pitches the root about its own lateral axis, with a term that keeps the feet at their original heights so that a lower root comes with bent legs (weight 100, same cost, regularizer, smoothness, learning rate and budget otherwise; closed loop every 5 ticks with 30 Adam steps, and every tick or with 100 steps as controls). MimicLite observes the root height and the root orientation of the reference, so the teacher sees the edit. The edited references clear the bar (mean worst violation after editing at most 1.0\,cm; 0.6\,cm offline and 0.3\,cm in closed loop at 1.25\,m) and lower the pelvis by up to 14\,cm, yet clean passes stay at 0.55, 0.03 and 0.00 offline and 0.67, 0.04 and 0.02 in closed loop for the 1.30, 1.25 and 1.20\,m bars (closed loop at 1.25\,m: 0.04 every tick, 0.02 with 100 steps). At the tick where the offline edit is deepest the teacher's pelvis is 0.2, 1.0 and 1.2\,cm below the original reference while the edit asks for 2.3, 6.2 and 9.3\,cm.
Reach steering also moves the root, so the same root editor is run on the reaching cost, with the legs freed in addition to the right arm and waist (offline; closed loop every tick as for the J editor). It hits the target in 0.31 (offline) and 0.60 (closed loop) of the trials, against 0.38 and 0.60 without the root. Hurdle steering keeps the root, so the J editor is already the matching control there.

\paragraph{A stronger tracker of edited references.}
To test whether the edits fail because of the teacher, we let SONIC (release), the tracker with the highest success in \cref{tab:track}, execute the same edited references in the MuJoCo harness of \cref{tab:track}. The harness has no obstacles, so we score the executed whole-body capsules against the obstacle geometry (a pass is clean if no capsule enters the bar by 1\,cm or more); without contact this is an upper bound. SONIC executes the edits no better than the teacher. Under the 1.25\,m bar the edited reference is clean in every window, but the executed motion is clean in 0.03 of them for SONIC and 0.11 for the teacher (0.06 and 0.09 with the root edit; of the 6.1\,cm by which this edit lowers the pelvis near the bar, the largest drop within 0.3\,m of it averaged over windows, SONIC realizes 2.0\,cm and the teacher 1.7\,cm); at 1.30\,m both reach 0.89, and at 1.20\,m neither passes. Over the hurdles the edited reference avoids contact in 0.96 of the windows and neither tracker in any: both keep their feet near the original height (mean penetration 5.7 and 5.5\,cm, against 5.8\,cm on the original reference).

\section{Additional results}
\label{sec:app}
\paragraph{Tracking on shared clips.}
$E_l$ in \cref{tab:track} averages over each method's own successful clips. On the clips that every method tracks (134 of 150 AMASS and 457 of 512 BONES windows) the best method is unchanged (\cref{tab:track_shared}): the \method{} trained from scratch has the lowest error on both sets, and the paired bootstrap interval (1000 resamples over clips) of every other method's difference to it excludes zero, e.g.\ $+1.5$\,mm $[0.9, 2.0]$ for HEFT on BONES and $+3.7$\,mm $[2.4, 5.2]$ on AMASS.

\begin{table}[h]
\centering
\small
\setlength{\tabcolsep}{4pt}
\begin{tabular}{@{}lcc@{}}
\toprule
Method & AMASS (134) & BONES (457) \\
\midrule
SONIC (release) & 29.4 [27.3, 31.8] & 25.2 [24.2, 26.3] \\
SONIC (v1.1) & 27.5 [25.9, 29.6] & 23.8 [22.8, 25.0] \\
HoloMotion & 39.5 [37.0, 42.4] & 35.4 [34.1, 37.1] \\
HEFT (pmg) & 27.5 [25.4, 30.2] & 21.9 [21.2, 22.7] \\
Humanoid-GPT & 39.2 [34.8, 44.3] & 31.3 [29.3, 33.6] \\
MimicLite Huge (teacher) & 28.1 [25.9, 30.8] & 24.2 [23.3, 25.2] \\
\method{} (distilled) & 28.1 [25.9, 30.5] & 24.4 [23.6, 25.4] \\
\method{} (RL from scratch) & \textbf{23.8} [22.1, 25.9] & \textbf{20.5} [19.7, 21.3] \\
\bottomrule
\end{tabular}
\caption{\textbf{Local error $E_l$ (mm) on the clips all methods track}, with 95\% bootstrap intervals over clips.}
\label{tab:track_shared}
\end{table}

\paragraph{Training-free objectives.}
\cref{tab:more} lists the numbers summarized in \cref{fig:q3}; the VR numbers are in \cref{tab:vr}.

\paragraph{Planned and executed torque.}
The torque cost acts on the torque implied by the planned actions, while the robot executes the action of the head. To check how much of the planned reduction reaches the motor, we log the torque of the weak joint at every tick (\cref{tab:tq}; one run per setting, the settings of \cref{tab:more}). Steering lowers the torque the head commands, on average from 1.00 to 0.69 of the degraded limit for the knee and from 1.47 to 1.07 for the hip, and shortens the time the weak motor saturates (13.9 to 9.9\% and 24.1 to 19.7\% of the ticks). The executed reduction is smaller than the planned one, and after steering the planned and commanded torques of the current tick agree less (correlation 0.88 to 0.31 for the knee, 0.92 to 0.55 for the hip). This is expected from how the head is trained: it learns to execute plans of natural motions, with no task, fault or steered data (\cref{sec:app_method}), so a plan that shifts effort away from one joint lies partly outside what it has seen, and it follows that plan only in part while its feedback keeps the robot balanced. Steering still pays off at the task level (success 0.71 to 0.79 and 0.77 to 0.86 in these runs), and closing the remaining gap, \eg by also training the head on steered plans in a task-agnostic way, is left to future work.

\begin{table}[h]
\centering
\small
\setlength{\tabcolsep}{3pt}
\resizebox{\linewidth}{!}{%
\begin{tabular}{@{}llcccccc@{}}
\toprule
 & & & \multicolumn{2}{c}{Planned / limit} & Commanded & Saturated & Corr.\ planned \\
\cmidrule(lr){4-5}
Weak joint & & Success$\uparrow$ & 16 ticks & now & / limit & ticks (\%) & vs.\ commanded \\
\midrule
Left knee & unsteered & 0.71 & 0.85 & 0.66 & 1.00 & 13.9 & 0.88 \\
 & steered & 0.79 & 0.62 & 0.37 & 0.69 & 9.9 & 0.31 \\
\midrule
Left hip pitch & unsteered & 0.77 & 1.09 & 0.91 & 1.47 & 24.1 & 0.92 \\
 & steered & 0.86 & 0.64 & 0.48 & 1.07 & 19.7 & 0.55 \\
\bottomrule
\end{tabular}}
\caption{\textbf{Planned and executed torque of the weak joint} (15\% of its torque, 150 BONES windows, one run each). Torques are absolute values over the live ticks, divided by the degraded limit. Planned: implied by the planned actions and states (PD law), the maximum over the 16 ticks the cost uses, or the current tick. Commanded: the PD torque of the action the head executes, before the motor clips it at the limit. Saturated: ticks at which the weak motor delivers at least 95\% of its limit.}
\label{tab:tq}
\end{table}

\paragraph{Planned and executed clearance.}
On the obstacles, we compare the clearance the executed plan predicts with the clearance the robot then has (\cref{tab:cl}; real-time setting, one run per setting, ticks at which the body is within 15\,cm of an obstacle). For the next tick, plan and execution agree within about 2\,cm (median absolute error) with and without steering. Further ahead the steered plan expects 3 to 10\,cm more clearance at 0.1 to 0.2\,s than the robot then has. The plan is regenerated from the measured state at every tick, so the far end of each plan is never executed as such; under the low bar the plan predicts a clear pass at a tick where the robot then touches the bar in 0.2\% of the cases, against 6 to 18\% without steering. On the hurdles this rate stays at 13 to 16\% with and without steering, which matches the foot contacts that feasible passes allow.

\begin{table}[h]
\centering
\small
\setlength{\tabcolsep}{3pt}
\resizebox{\linewidth}{!}{%
\begin{tabular}{@{}llccc@{}}
\toprule
 & & \multicolumn{3}{c}{Plan horizon} \\
\cmidrule(lr){3-5}
Task & & 1 tick & 5 ticks (0.1\,s) & 10 ticks (0.2\,s) \\
\midrule
Low bar 1.25\,m & unsteered & $-1.2$ / 1.5 / 5.6 & $-2.2$ / 2.7 / 7.8 & $-1.5$ / 3.6 / 17.7 \\
 & steered & 1.3 / 2.0 / 0.2 & 3.9 / 4.0 / 0.2 & 9.6 / 9.7 / 0.2 \\
\midrule
Hurdles 10\,cm & unsteered & $-1.0$ / 2.9 / 11.8 & $-1.1$ / 3.0 / 14.3 & $-1.2$ / 3.5 / 14.3 \\
 & steered & 1.2 / 2.3 / 13.1 & 2.8 / 3.5 / 14.8 & 5.5 / 5.8 / 16.2 \\
\bottomrule
\end{tabular}}
\caption{\textbf{Planned vs.\ executed whole-body clearance} (79 walking windows, one run each, ticks with the body within 15\,cm of an obstacle). Each cell: median of planned minus executed clearance (cm) / median absolute difference (cm) / ticks at which the plan predicts no contact but the robot touches (\%). Clearance: minimum over the collision-capsule sample points of the obstacle distance minus the capsule radius, the same geometry as the cost.}
\label{tab:cl}
\end{table}

\begin{table}[h]
\centering
\small
\setlength{\tabcolsep}{3pt}
\resizebox{\linewidth}{!}{%
\begin{tabular}{@{}llccc@{}}
\toprule
Task & Metric & Teacher & \method{} & \method{} steered \\
\midrule
Left knee at 15\% torque & success rate$\uparrow$ & 0.69 & 0.71 & \textbf{0.79} \\
Left hip pitch at 15\% & success rate$\uparrow$ & 0.78 & 0.79 & \textbf{0.84} \\
\midrule
Tray balancing (4\,cm rim) & payload lost rate$\downarrow$ & 0.18 & 0.16 & \textbf{0.08} \\
\midrule
Velocity command tracking & velocity error (m/s)$\downarrow$ & 0.157 & 0.159 & \textbf{0.112} \\
Spatial reaching (12 steps) & target hit rate$\uparrow$ & 0.01 & 0.02 & \textbf{0.75} \\
\bottomrule
\end{tabular}}
\caption{\textbf{Training-free adaptation across diverse physical objectives}. Evaluated in simulation; ``\method{}'' represents nominal unsteered execution. Steered: both generator calls, two steps each, except reaching (12 smaller steps).}
\label{tab:more}
\end{table}

\paragraph{Steering the model trained from scratch.}
\label{sec:app_rl}
With the same costs and settings (both calls steered, two steps each), the model trained from scratch gains from steering on every task (\cref{tab:rl}): feasible hurdle passes rise from 0.13 to 0.37 and high-five hits from 0.00 to 0.46, and it passes under the 1.25 and 1.20\,m bars more often than the distilled model (0.93 and 0.72 against 0.86 and 0.53). It gains less on hurdles and reaching, and the distilled action head on its plan recovers only a small part of the gap (0.43 and 0.56), so on these two tasks mainly its plan limits it. The steering experiments of the main paper use the distilled model because it shares the teacher's training data, simulator and reference interface, so steering inside the generator and editing the teacher's reference are compared in the same setting.

\begin{table}[h]
\centering
\small
\setlength{\tabcolsep}{3pt}
\resizebox{\linewidth}{!}{%
\newcommand{\sd}[1]{{\scriptsize${\pm}$#1}}
\begin{tabular}{@{}lccccc@{}}
\toprule
 & & & & \multicolumn{2}{c}{Low bar clean} \\
\cmidrule(l){5-6}
Training & GPU-h & Hurdles & High-five & 1.25\,m & 1.20\,m \\
\midrule
Distilled from teacher & 32 & \textbf{0.70}\sd{0.02} & \textbf{0.57}\sd{0.04} & 0.86\sd{0.02} & 0.53\sd{0.04} \\
RL from scratch & \textbf{20} & 0.37\sd{0.06} & 0.46\sd{0.03} & \textbf{0.93}\sd{0.02} & \textbf{0.72}\sd{0.02} \\
RL plan + distilled head & -- & 0.43\sd{0.04} & 0.56\sd{0.02} & 0.90\sd{0.02} & 0.68\sd{0.04} \\
\bottomrule
\end{tabular}}
\caption{\textbf{Steering the model trained without a teacher} (the 60k-iteration model of \cref{tab:track}; training simulator, 79 walking windows, both calls steered as in the both-calls row of \cref{tab:steer}, mean${\pm}$s.d.\ over three runs; the distilled model was re-run in the same batch). Hurdles: feasible passes; high-five: hits within 8\,cm. GPU-h: GPU-hours on RTX 4090 GPUs (four per run), same motion library.}
\label{tab:rl}
\end{table}

\paragraph{Impulse pushes.}
Root-velocity kicks every 2--3\,s at several multiples of the training perturbation (\cref{tab:push}); \method{} and its teacher are indistinguishable.

\begin{table}[h]
\centering
\small
\setlength{\tabcolsep}{4pt}
\begin{tabular}{@{}llcccc@{}}
\toprule
 & & \multicolumn{4}{c}{Push scale} \\
\cmidrule(l){3-6}
Metric & Method & 0 & 2 & 4 & 6 \\
\midrule
\multirow{2}{*}{Succ.\,(\%)$\uparrow$} & Teacher & \textbf{95.5} & 95.6 & 95.1 & \textbf{94.8} \\
 & \method{} & \textbf{95.5} & \textbf{95.9} & \textbf{95.5} & 94.6 \\
\midrule
\multirow{2}{*}{$E_l$\,(mm)$\downarrow$} & Teacher & \textbf{28.8} & 29.8 & 31.7 & 33.6 \\
 & \method{} & 29.0 & \textbf{29.7} & \textbf{31.4} & \textbf{33.5} \\
\bottomrule
\end{tabular}
\caption{\textbf{Impulse pushes} on 512 unseen windows. Root-velocity kicks every 2--3\,s, with the scale given relative to the training perturbation. \method{} and its teacher are indistinguishable.}
\label{tab:push}
\end{table}

\paragraph{Latency.}
\label{sec:app_latency}
\cref{tab:latency} separates the computation of each setting. The generation step is one forward pass of the generator; a steering update is one forward and one backward pass through it, so the cost grows with the number of gradient steps, not with a denoising chain. On the RTX 4090 the whole steered step is captured as one CUDA graph (batch of one). On the Orin NX (MAXN) the step is captured in the same way and timed back to back over 500 steps, including the input and output checks; the steering cost is compiled with torch.compile (FP32; actions within $10^{-6}$\,rad of the uncompiled step), and for both calls the generator and action head are compiled as well, with TF32 matrix products (actions within 0.006\,rad; in simulation this leaves steering unchanged: hurdles $0.68{\pm}0.03$ vs.\ $0.66{\pm}0.03$, 1.25\,m bar $0.84{\pm}0.01$ vs.\ $0.83{\pm}0.04$, reach $0.54{\pm}0.05$ vs.\ $0.52{\pm}0.03$, three runs each). Uncompiled, the two steered settings take 28.3 and 48.7\,ms. The deployed ONNX tracker, timed when it runs once every 20\,ms as in deployment, takes 13.1\,ms (a full control step including observation assembly and command output takes 15.5\,ms); it is not captured as a graph, so these paced timings are not comparable with the back-to-back ones of \cref{tab:latency}, and we did not time steering in the paced loop. In the hardware trials steering ran on a tethered RTX 4090 (below).
Only the settings whose 95th percentile stays below 20\,ms on the Orin NX fit the synchronous 50\,Hz loop of the robot; the larger-budget settings and world-model projection do not, and we report them as computation--performance trade-offs rather than as real-time settings.

\begin{table}[h]
\centering
\small
\setlength{\tabcolsep}{3pt}
\resizebox{\linewidth}{!}{%
\begin{tabular}{@{}lccc@{}}
\toprule
 & & RTX 4090 & Orin NX \\
Setting & Gradient steps & median / p95 & median / p95 \\
\midrule
Tracking, 1 call & 0 & 3.8 / -- & 2.4 / 2.4 \\
Tracking, 2 calls (as on the robot) & 0 & 5.3 / 5.8 & 4.2 / 4.2 \\
\midrule
Steered, 2nd call only (as in the robot runs) & 2 & 10.6--13.7 / 15.3 & 14.4 / 14.6 \\
Steered, both calls & 2 + 2 & 17.2--20.5 / 20.7 & 18.2 / 18.4 \\
\midrule
Larger budget, small steps & 12 + 12 & 82.4--91.4 / 91.7 & 105.2 / 105.4 \\
Larger budget, line search & 6 + 6 & 100.4--111.7 / 112.1 & -- \\
\bottomrule
\end{tabular}}
\caption{\textbf{Computation per control step} (ms; policy inference only, batch of one). A range spans the steering costs (hurdles and reaching; also the low bar for the larger-budget rows; two runs each); on the Orin NX, hurdles and low bar agree within 0.1\,ms. The line search is timed with all four halvings run, an upper bound. Orin NX: compiled steering cost, and for both calls a compiled generator with TF32 (see text). ``--'': not measured (on the Orin NX the line search is not captured as a CUDA graph); the larger-budget settings exceed the 20\,ms budget on either device.}
\label{tab:latency}
\end{table}

\paragraph{Steering on the robot.}
\label{sec:app_real}
\begin{figure*}[t]
  \centering
  \includegraphics[width=0.98\textwidth]{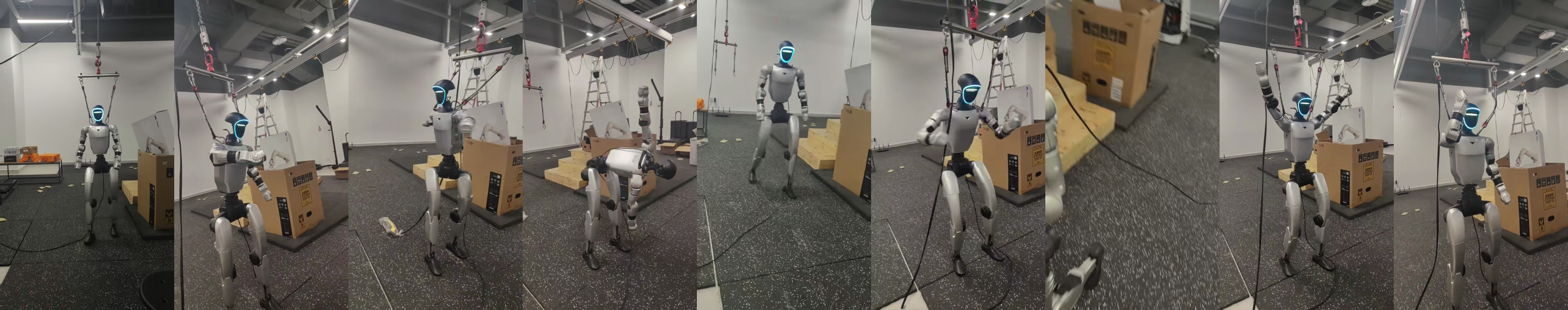}
  \caption{\textbf{Nominal tracking on a physical Unitree G1.}
  Representative frames from nine whole-body motion sequences.
  The overhead tether is an unweighted safety guard.}
  \label{fig:real}
\end{figure*}

\begin{figure*}[t]
  \centering
  \includegraphics[width=0.98\textwidth]{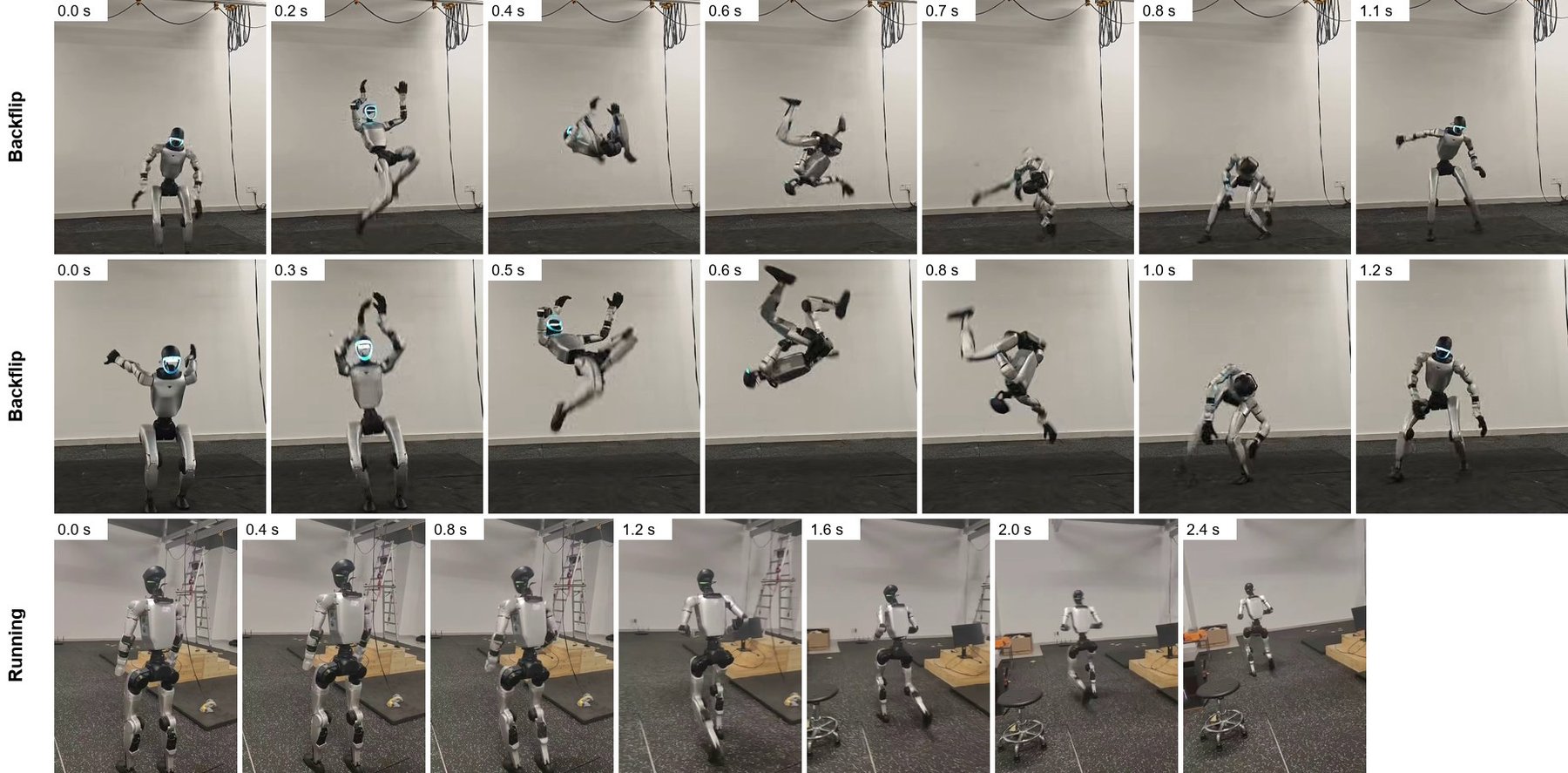}
  \caption{\textbf{Highly dynamic motions on a physical Unitree G1}, tracked by \method{} without steering: two backflips and a run (time from the first frame of each row).}
  \label{fig:real_dyn}
\end{figure*}

\begin{figure*}[t]
  \centering
  \includegraphics[width=0.8\textwidth]{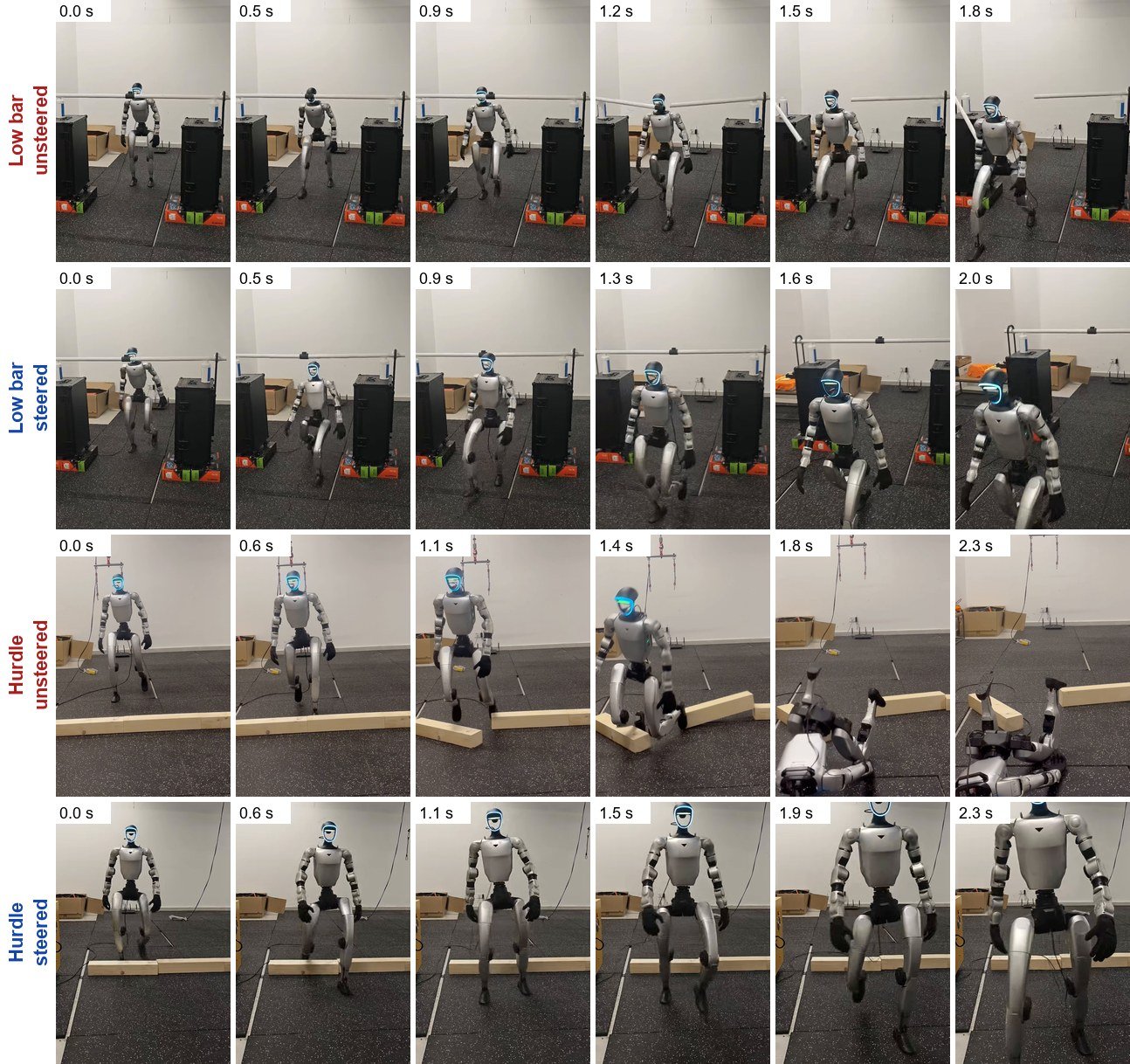}
  \caption{\textbf{Steering on the robot over time} (same frozen policy and walking reference in all rows; LiDAR distance field). Low bar at 1.1\,m: unsteered, the head hits the bar and knocks it off; steered, the robot lowers and passes under. Hurdle of 12\,cm: unsteered, the foot catches the hurdle and the robot falls; steered, the robot raises its feet and steps over.}
  \label{fig:real_seq}
\end{figure*}

The real-robot steering runs use the distilled \method{} (the same weights as in simulation) on a walking reference, with guidance in the second generator call only (two gradient steps). The cost is the whole-body SDF cost of the simulated low-bar task, evaluated on a distance field from the head-mounted LiDAR (Livox MID-360) instead of known obstacle geometry. Before a run, a short LiDAR capture registered by LiDAR-inertial odometry (FAST-LIO2) is turned into an occupancy map and a 3D Euclidean distance field (ROG-Map, 1\,cm voxels), which is then kept fixed; during the run the odometry only updates the robot's pose in this map, and unobserved voxels add no cost. In these runs the generator, the cost and the action head run on a tethered computer with an RTX 4090, which also holds the distance field; the Orin NX runs the LiDAR-inertial odometry.
Under a bar at 1.1\,m, the robot passed under the bar in four of five attempts; in the remaining attempt the robot crouched too far and its waist touched the ground. Without steering, the robot walked into the bar in all three attempts and broke it.
Over a 12\,cm hurdle, with the same cost, map pipeline and settings, the robot stepped over the hurdle in all three attempts: one crossing was clean, in one the foot landed on top of the hurdle and in one the foot kicked it. Without steering, the robot tripped over the hurdle in all three attempts.
\cref{fig:real_seq} shows one attempt of each setting as an image sequence, and \cref{fig:real_dyn} shows highly dynamic motions (backflips, running) tracked on the same robot without steering.

\paragraph{Map, pose drift and obstacle height in simulation.}
On the robot the cost reads a voxel distance field through an odometry pose. We reproduce this in simulation (\cref{tab:maperr}, real-time setting, three runs): the analytic SDF is replaced by the same voxel pipeline (3\,cm voxels) built from the true obstacles, and the pose is perturbed by an SE(2) random walk whose s.d.\ grows with the square root of the distance walked. The clean pass rate under the 1.25\,m bar degrades gracefully with the drift and stays far above the unsteered 0.02 even at 11\,cm and 2.1$^\circ$ of drift at the bar; on hurdles, steering keeps most of its gain at every level. Steering also helps across hurdle heights: at 6\,cm it raises feasible passes from $0.75{\pm}0.02$ to $0.88{\pm}0.02$, and at 14\,cm, where the unsteered tracker mostly falls, it raises success from $0.23{\pm}0.03$ to $0.59{\pm}0.04$.

\begin{table}[h]
\centering
\small
\setlength{\tabcolsep}{4pt}
\newcommand{\sd}[1]{{\scriptsize${\pm}$#1}}
\begin{tabular}{@{}lcc@{}}
\toprule
Cost evaluated on & Low bar 1.25\,m$\uparrow$ & Hurdles$\uparrow$ \\
\midrule
-- (unsteered) & 0.02 & 0.40 \\
Analytic SDF (\cref{tab:steer}) & 0.90\sd{0.01} & 0.68\sd{0.04} \\
Voxel map, exact pose & 0.77\sd{0.06} & 0.65\sd{0.02} \\
\quad drift 3.7\,cm / 0.7$^\circ$ & 0.55\sd{0.08} & 0.58\sd{0.06} \\
\quad drift 7.3\,cm / 1.4$^\circ$ & 0.35\sd{0.07} & 0.52\sd{0.03} \\
\quad drift 11\,cm / 2.1$^\circ$ & 0.21\sd{0.08} & 0.48\sd{0.05} \\
\bottomrule
\end{tabular}
\caption{\textbf{Map and pose errors} (79 walking windows; clean passes under the bar, feasible passes over 10\,cm hurdles; mean${\pm}$s.d.\ over three runs). Drift: s.d.\ of the position and heading error after 1.5\,m, the distance to the bar.}
\label{tab:maperr}
\end{table}

\paragraph{VR three-point teleoperation.}
With only the root reference visible (legs and upper body masked), a cost on the head and both wrists drives \method{} from a headset and two controllers; none of this input was seen in training. \cref{tab:vr} (150 unseen windows, one run each) shows that guidance brings the wrist error from about 24 to 16.4\,cm; with world-model projection the run gives 14.5\,cm, a difference a single run does not establish. A root estimated from the headset raises the fall rate from 2.7\% (full reference) to 9.3\% because of its posture; giving the waist height, as the interface of SONIC does, brings it to 3.3\%. The wrist position relative to the head is 12.9\,cm (11.6\,cm with projection; 4.3\,cm with full tracking), so this is usable for coarse teleoperation only.

\begin{table}[h]
\centering
\small
\setlength{\tabcolsep}{3pt}
\resizebox{\linewidth}{!}{%
\begin{tabular}{@{}lcccc@{}}
\toprule
Setting & fall & head (cm) & wrist (cm) & wrist rel.\ head (cm) \\
\midrule
Full reference & 0.027 & 10.2 & 10.8 & 4.3 \\
Root from headset, no guidance & 0.093 & 13.8 & 24.7 & 19.6 \\
Waist height given, no guidance & 0.033 & 14.0 & 24.3 & 19.4 \\
Waist height given, guidance 0.5 & 0.033 & 12.1 & 16.4 & 12.9 \\
\quad + world-model projection & 0.033 & 12.9 & 14.5 & 11.6 \\
\bottomrule
\end{tabular}}
\caption{\textbf{VR three-point targets} (one run each).}
\label{tab:vr}
\end{table}

\paragraph{No-step zones and dodging.}
A cost that keeps the planned stance foot out of a zone on the reference foothold lowers the fraction of feet landing in it from 0.62 to 0.21 (strength 2) with unchanged success (0.953). For balls thrown at the robot along a known path (4\,m/s, two per clip), core hits fall from 0.94 to 0.60 at strength 8 with no extra falls but doubled tracking error; with only the waist and arms free the reduction in hits over decoy balls is 0.18.

\paragraph{All steering metrics.}
\cref{tab:steer_full} extends \cref{tab:steer} with severe failures (body contact or fall at a hurdle), overall success and the closest wrist distance.
Under the 1.20\,m bar, guidance in the second call passes cleanly in 0.56 of the windows. Recomputing the contacts from the executed poses (in agreement with the simulator's contact flag in 61 of 66 windows per run), every contact starts at the head (26--38\% of the windows per run), never at the arms or legs. The robot does duck, with its pelvis 6--8\,cm lower at the bar than at the start of the window, but not far enough to bring the head under the bar, in line with the partial execution of guided plans (\cref{sec:steps}).

\begin{table*}[h]
\centering
\small
\resizebox{\linewidth}{!}{%
\begin{tabular}{@{}lcccccccccc@{}}
\toprule
 & \multicolumn{4}{c}{Hurdles} & \multicolumn{3}{c}{Low bar: clean pass\,$\uparrow$} & \multicolumn{2}{c}{Reach (high-five)} \\
\cmidrule(lr){2-5}\cmidrule(lr){6-8}\cmidrule(lr){9-10}
Method & feasible$\uparrow$ & severe$\downarrow$ & clean$\uparrow$ & succ.$\uparrow$ & 1.30\,m & 1.25\,m & 1.20\,m & hit$\uparrow$ & dist.\,(cm)$\downarrow$ \\
\midrule
Teacher, original reference & 0.48 & 0.33 & 0.01 & 0.63 & 0.36 & 0.02 & 0.00 & 0.01 & 20 \\
Teacher, offline-edited reference & 0.58 & 0.29 & 0.01 & 0.78 & 0.72 & 0.06 & 0.00 & 0.38 & 12 \\
Teacher, closed-loop edited reference & 0.57 & 0.25 & 0.02 & 0.78 & 0.87 & 0.11 & 0.02 & 0.60 & 8 \\
Teacher, offline edit with root & 0.47 & 0.36 & 0.00 & 0.67 & 0.55 & 0.03 & 0.00 & 0.31 & 13 \\
Teacher, closed-loop edit with root & 0.45 & 0.33 & 0.00 & 0.69 & 0.67 & 0.04 & 0.02 & 0.60 & 9 \\
\method{}, no steering & 0.40 & 0.37 & 0.00 & 0.67 & 0.23 & 0.02 & 0.00 & 0.02 & 21 \\
\method{}, steered (2nd call only: 2 steps) & 0.68 & 0.20 & 0.11 & 0.95 & 0.95 & \textbf{0.90} & \textbf{0.56} & 0.57 & 9 \\
\method{}, steered (both calls: 2 steps per call) & 0.66 & 0.19 & 0.12 & 0.92 & 0.95 & 0.84 & 0.52 & 0.54 & 9 \\
\method{}, steered (larger budget: 12 steps at 1/8 strength) & 0.70 & 0.19 & 0.03 & \textbf{0.96} & 0.88 & 0.22 & 0.03 & \textbf{0.75} & \textbf{7} \\
\method{}, steered (larger budget: 6 steps + line search) & \textbf{0.74} & \textbf{0.16} & \textbf{0.21} & 0.95 & \textbf{0.98} & 0.83 & \textbf{0.56} & 0.62 & 8 \\
\bottomrule
\end{tabular}}
\caption{\textbf{Steering vs.\ editing the reference, all metrics} (79 walking windows; steered rows are the mean of three runs, the controls of two to three; ``no steering'' sets the strength to zero). The teacher tracks a reference that was edited with the same cost, offline or in closed loop (re-optimized from the robot's root at every tick, every 5 ticks for the low bar); \method{} is steered inside the generator and tracks the original reference. Hurdles: clean is a crossing without contact, feasible also counts contact by the feet only, severe is body contact or a fall. Each larger-budget row applies one procedure to all three tasks (three runs each); see \cref{sec:steps}. Over hurdles the root editors let the feet rise but not sink (a one-sided foot term; with no foot term, 0.44 offline and 0.43 in closed loop); the latency of offline edits excludes the edit, which is computed before the rollout; closed-loop edits are timed when re-optimized at every tick; the line search is timed with all halvings run.}
\label{tab:steer_full}
\end{table*}

\begin{table}[h]
\centering
\small
\setlength{\tabcolsep}{2pt}
\resizebox{\linewidth}{!}{%
\newcommand{\sd}[1]{{\scriptsize${\pm}$#1}}
\begin{tabular}{@{}lccccc@{}}
\toprule
 & & \multicolumn{3}{c}{Low bar clean} & \\
\cmidrule(lr){3-5}
 & Hurdles & 1.30\,m & 1.25\,m & 1.20\,m & Reach \\
\midrule
Teacher, original ref. & 0.48\sd{0.02} & 0.36\sd{0.00}$^2$ & 0.02\sd{0.00}$^2$ & 0.00\sd{0.00}$^2$ & 0.01\sd{0.01}$^2$ \\
Teacher, offline edit (J) & 0.58\sd{0.02} & 0.72\sd{0.01}$^2$ & 0.06\sd{0.00}$^2$ & 0.00\sd{0.00}$^2$ & 0.38\sd{0.01} \\
Teacher, offline edit (J+R) & 0.47\sd{0.05} & 0.55\sd{0.05} & 0.03\sd{0.00} & 0.00\sd{0.00} & 0.31\sd{0.02} \\
Teacher, closed-loop edit (J) & 0.57\sd{0.04} & 0.87\sd{0.02} & 0.11\sd{0.00} & 0.02\sd{0.00} & 0.60\sd{0.01} \\
Teacher, closed-loop edit (J+R) & 0.45\sd{0.09} & 0.67\sd{0.05} & 0.04\sd{0.01} & 0.02\sd{0.00} & 0.60\sd{0.01} \\
\method{}, unsteered & 0.40$^1$ & 0.23$^1$ & 0.02$^1$ & 0.00$^1$ & 0.02\sd{0.01}$^2$ \\
\method{}, 2nd call, 2 steps & 0.68\sd{0.04} & 0.95\sd{0.02} & 0.90\sd{0.01} & 0.56\sd{0.04} & 0.57\sd{0.03} \\
\method{}, both calls, 4 steps & 0.66\sd{0.03} & 0.95\sd{0.01} & 0.84\sd{0.02} & 0.52\sd{0.04} & 0.54\sd{0.06} \\
\method{}, 6 steps + line search & 0.74\sd{0.01} & 0.98\sd{0.01} & 0.83\sd{0.04} & 0.56\sd{0.01} & 0.62\sd{0.02} \\
\bottomrule
\end{tabular}}
\caption{\textbf{Spread of \cref{tab:steer}} (mean${\pm}$s.d.\ over three runs; superscripts mark one or two runs).}
\label{tab:steer_sd}
\end{table}

\paragraph{Window-level uncertainty.}
With 79 windows a success rate has a binomial standard error of about 0.05, so we also compare the real-time row of \cref{tab:steer} with each editor window by window (\cref{tab:paired}): per window, the share of its bars passed, averaged over the runs; a paired bootstrap over windows (10{,}000 resamples) and a sign test over the windows whose results differ. Under the 1.25 and 1.20\,m bars the gain over every editor is far outside the uncertainty. On hurdles the gain over the offline editor is significant, but over the closed-loop editor the interval includes zero, and at 1.30\,m the gain over it is small; we therefore claim a wide margin only under the lower bars.

\begin{table}[h]
\centering
\small
\setlength{\tabcolsep}{3pt}
\resizebox{\linewidth}{!}{%
\begin{tabular}{@{}llccc@{}}
\toprule
Task & Editor (teacher) & Difference & 95\% interval & Sign test $p$ \\
\midrule
Hurdles, feasible & offline edit (J) & $+0.11$ & $[+0.02, +0.19]$ & 0.03 \\
 & closed-loop edit (J) & $+0.06$ & $[-0.03, +0.15]$ & 0.03 \\
Low bar 1.30\,m, clean & offline edit (J) & $+0.24$ & $[+0.12, +0.35]$ & $9{\cdot}10^{-4}$ \\
 & closed-loop edit (J) & $+0.08$ & $[+0.01, +0.17]$ & 0.15 \\
Low bar 1.25\,m, clean & closed-loop edit (J) & $+0.80$ & $[+0.70, +0.89]$ & $2{\cdot}10^{-15}$ \\
Low bar 1.20\,m, clean & closed-loop edit (J) & $+0.54$ & $[+0.44, +0.64]$ & $3{\cdot}10^{-14}$ \\
\bottomrule
\end{tabular}}
\caption{\textbf{Window-level comparison} of \method{} (real-time setting) with the strongest reference editors of \cref{tab:steer} (79 walking windows, those with an active obstacle: 73 for hurdles, 66 for the low bar; three runs, two for the teacher's offline low-bar rows). Difference: \method{} minus editor in the per-window pass rate; interval: paired bootstrap over windows. Reaching is not included because its runs store no per-window hits.}
\label{tab:paired}
\end{table}

\paragraph{Body-group masking.}
\label{sec:app_mask}
\cref{tab:mask} gives the full results of masking the reference for the generator and the executor (\cref{sec:more}). The obstacle rows of \cref{tab:steer} hide the legs from the executor only; hiding them from the generator as well leaves the obstacles unchanged (hurdles 0.66 vs.\ 0.68 feasible, success 0.94 vs.\ 0.91; 1.25\,m bar 0.84 clean in both; three runs, both calls steered).
The weak-motor costs show the same ordering as the velocity command at a smaller scale across four faults: masking the generator alone does not help (0.62--0.83 against 0.63--0.84 with the reference visible), while hiding the leg reference from the executor too raises success on the three knee faults by 0.03--0.05 in both of its runs (e.g., 0.82 to 0.86 for both knees at 20\%) and on the hip fault by only 0.01, within the run-to-run spread, at the price of a higher tracking error ($+$13 to $+$15\,mm), because the legs no longer follow the reference.

\begin{table}[t]
\centering
\small
\setlength{\tabcolsep}{3pt}
\resizebox{\linewidth}{!}{%
\begin{tabular}{@{}lccc@{}}
\toprule
 & \multicolumn{3}{c}{Reference hidden from} \\
\cmidrule(l){2-4}
Task where the reference conflicts with the cost & none & generator & generator + executor \\
\midrule
Velocity command, unguided: velocity error (m/s)$\downarrow$ & 0.218 & 0.205 & \textbf{0.159} \\
Velocity command, unguided: heading error ($^\circ$)$\downarrow$ & 14.7 & 14.1 & \textbf{6.2} \\
Velocity command, root-position cost: velocity error (m/s)$\downarrow$ & 0.173 & 0.168 & \textbf{0.139} \\
Velocity command, root-position cost: heading error ($^\circ$)$\downarrow$ & 8.2 & 7.3 & \textbf{2.4} \\
Left knee at 15\%, guided: success rate$\uparrow$ & 0.79 & 0.79 & \textbf{0.82} \\
Both knees at 20\%, guided: success rate$\uparrow$ & 0.82 & 0.81 & \textbf{0.86} \\
Left hip pitch at 15\%, guided: success rate$\uparrow$ & 0.84 & 0.83 & \textbf{0.85} \\
Left knee at 10\%, guided: success rate$\uparrow$ & 0.63 & 0.62 & \textbf{0.67} \\
\midrule
\multicolumn{4}{@{}l}{Tracking under sparse references, no cost (512 unseen windows)} \\
Reference masked for & success rate$\uparrow$ & $E_l$\,(mm)$\downarrow$ & \\
nominal (none) & 0.955 & 29.1 & \\
legs masked & 0.955 & 40.5 & \\
upper body masked & 0.959 & 43.5 & \\
root and legs masked (upper body only) & 0.926 & 48.2 & \\
\bottomrule
\end{tabular}}
\caption{\textbf{Body-group masking and handover.} Top: when the reference conflicts with the cost, hiding it from the generator alone is not enough; the executor must also stop seeing it. The velocity command hides the leg and upper-body references and keeps the root path; the knee fault hides the leg reference; fault rows: masked settings mean of two runs, reference visible one run (left knee at 15\%: three runs). Bottom: sparse tracking preserves stability under partial references.}
\label{tab:mask}
\end{table}

\paragraph{What the relaxed and the unrelaxed parts of the reference do.}
\label{sec:app_retain}
Hiding a group hands it to the cost, but the other groups should still follow the reference. \cref{tab:retain} measures both for reaching and the tray, whose costs act on the right arm, so the upper body is the relaxed group. Each setting uses the same frozen weights, two generator calls, both steered (two gradient steps per call; strength 2 for reaching, 0.5 for the tray). We report the task metric over all clips and, on the clips that succeed in every setting, the mean absolute joint error to the original reference per group, the root height and heading errors, and the horizontal root drift at the end of the window. Steering leaves the unrelaxed groups almost unchanged: the leg error moves from 4.4$^\circ$ to at most 4.5$^\circ$, the root height error from 0.74 to at most 0.77\,cm and the heading error by at most 0.2$^\circ$; the drift grows from 21 to 22--25\,cm. The cost is paid by the relaxed group, including the left arm, which belongs to it although the cost does not use it. Where the mask is applied decides the trade-off. Hiding the arm reference from the executor only (our default) keeps the left arm within 2.8$^\circ$ of nominal tracking and gives the lowest ball loss (0.07, mean of three runs); hiding it from the generator as well gives the most reaching hits (0.65) but lets the left arm drift by 8$^\circ$; hiding it from the generator only behaves like no mask, since the executor still pulls the arm back to the reference. On 512 held-out windows of all motion types (standing, turning, upper-body motions), with the same strength and no tuning, the unrelaxed groups change as little (legs 4.6 to 4.8$^\circ$, root height 1.21 to 1.27\,cm, heading 2.6 to 3.0$^\circ$) while the hit rate rises from 0.02 to 0.52 (executor mask) or 0.55 (both masks) at a success rate of 0.95 against 0.96. Raising the strength to 4 adds no hits and raises the drift to 28--34\,cm, so the default strength sits at the useful end of the trade-off.

\begin{table*}[t]
\centering
\small
\setlength{\tabcolsep}{4pt}
\resizebox{\linewidth}{!}{%
\begin{tabular}{@{}llcccccccc@{}}
\toprule
 & Arm reference hidden from & Task$\uparrow$/$\downarrow$ & Legs ($^\circ$) & Waist ($^\circ$) & Left arm ($^\circ$) & Right arm ($^\circ$) & Root height (cm) & Heading ($^\circ$) & Drift (cm) \\
\midrule
\multirow{6}{*}{\shortstack[l]{Reaching\\(hit rate$\uparrow$)}} & none, no steering & 0.00 & 4.40 & 2.21 & 2.92 & 3.27 & 0.74 & 2.39 & 20.7 \\
 & none & 0.34 & 4.49 & 2.24 & 3.46 & 4.13 & 0.76 & 2.57 & 22.4 \\
 & generator & 0.29 & 4.49 & 2.25 & 3.48 & 4.26 & 0.76 & 2.58 & 22.3 \\
 & executor (default) & 0.53 & 4.53 & 2.19 & 5.66 & 7.27 & 0.77 & 2.53 & 22.2 \\
 & generator + executor, no steering & 0.03 & 4.38 & 2.12 & 7.82 & 9.78 & 0.74 & 2.28 & 19.0 \\
 & generator + executor & \textbf{0.65} & 4.54 & 2.23 & 10.81 & 14.79 & 0.77 & 2.59 & 24.8 \\
\midrule
\multirow{6}{*}{\shortstack[l]{Tray, 4\,cm rim\\(ball lost$\downarrow$)}} & none, no steering & 0.16 & 4.40 & 2.24 & 2.92 & 3.34 & 0.74 & 2.35 & 20.4 \\
 & none & 0.15 & 4.41 & 2.29 & 2.94 & 3.49 & 0.74 & 2.47 & 20.9 \\
 & generator & 0.17 & 4.40 & 2.25 & 3.05 & 3.54 & 0.74 & 2.44 & 20.6 \\
 & executor (default) & \textbf{0.07} & 4.42 & 2.18 & 4.67 & 6.29 & 0.75 & 2.32 & 20.4 \\
 & generator + executor, no steering & 0.17 & 4.35 & 2.14 & 7.88 & 9.93 & 0.73 & 2.22 & 18.9 \\
 & generator + executor & 0.11 & 4.41 & 2.15 & 7.84 & 10.18 & 0.75 & 2.33 & 19.6 \\
\bottomrule
\end{tabular}}
\caption{\textbf{Task gain against tracking of the reference that is not relaxed} (79 walking windows, mean of three runs, all windows succeed in every setting). The tray rows are a separate batch of three runs; the both-calls tray result of \cref{tab:calls} (0.08) comes from another batch with the same setting. Errors are to the original reference; legs, waist, root height and heading are not relaxed, the arms are. Steered rows use the default strength.}
\label{tab:retain}
\end{table*}

\paragraph{Retained groups under obstacle steering, against the baselines.}
\label{sec:app_retain_bar}
\cref{tab:retain_bar} repeats the measurement for the ceiling bar, where the cost acts on the legs and waist and the arms are pinned to the reference, and adds the unsteered teacher, which tracks the reference but cannot take a new cost at test time, so it only gives the tracking level to compare against. Steering leaves the retained groups where unsteered tracking has them: the arm error never rises (it falls by up to 0.3$^\circ$), the heading error rises by at most 0.2$^\circ$ (and falls at 1.25 and 1.20\,m), and the root height by at most 0.3\,cm, with success 1.00 at all three heights. The legs are the relaxed group here, so their error is not a retained-group measure. The reaching and tray costs of \cref{tab:retain} do not involve the legs, and for them the retained groups change as little. Not every task is free of such a cost: when the cost moves the feet (a half-circle foot target), the heading error rises from 2.6 to 5.6$^\circ$, the root height from 0.8 to 2.9\,cm and success falls from 1.00 to 0.92, because the legs carry the root. The unsteered local error of \method{} is above the teacher's under the bars (29.3--44.9 vs.\ 24.5--36.0\,mm) because of the bar: on the same 79 windows without obstacles the two are on par (24.1 vs.\ 23.9\,mm, two runs each).

\begin{table}[t]
\centering
\small
\setlength{\tabcolsep}{3pt}
\resizebox{\linewidth}{!}{%
\begin{tabular}{@{}llcccccc@{}}
\toprule
Bar & Method & Arms ($^\circ$) & Heading ($^\circ$) & Height (cm) & Legs$^{\dagger}$ ($^\circ$) & $E_l$ (mm) & Success \\
\midrule
\multirow{3}{*}{1.30\,m} & teacher, unsteered & 3.5 & 2.3 & 0.8 & 3.7 & 24.5 & 1.00 \\
 & \method{}, unsteered & 3.1 & 2.6 & 0.9 & 4.8 & 29.3 & 1.00 \\
 & \method{}, steered & 3.1 & 2.8 & 0.9 & 4.9 & 29.9 & 1.00 \\
\midrule
\multirow{3}{*}{1.25\,m} & teacher, unsteered & 4.0 & 3.0 & 1.0 & 4.6 & 31.8 & 1.00 \\
 & \method{}, unsteered & 3.4 & 3.5 & 1.0 & 5.7 & 37.5 & 1.00 \\
 & \method{}, steered & 3.2 & 2.8 & 1.1 & 5.5 & 32.8 & 1.00 \\
\midrule
\multirow{3}{*}{1.20\,m} & teacher, unsteered & 4.4 & 3.9 & 1.2 & 5.1 & 36.0 & 1.00 \\
 & \method{}, unsteered & 3.7 & 4.1 & 1.3 & 6.5 & 44.9 & 1.00 \\
 & \method{}, steered & 3.4 & 3.3 & 1.6 & 6.3 & 36.8 & 1.00 \\
\bottomrule
\end{tabular}}
\caption{\textbf{Tracking of the retained groups under the ceiling bar} (79 walking windows; steered rows are the mean of three runs, the teacher of two, the unsteered \method{} of one). Errors are to the original reference. The arms are pinned and the heading and root height are not relaxed; $^{\dagger}$the legs are the relaxed group, shown for reference. The teacher cannot take the cost, so its row is the tracking level, not a solution of the task.}
\label{tab:retain_bar}
\end{table}

\begin{figure*}[t]
  \centering
  \includegraphics[width=0.98\textwidth]{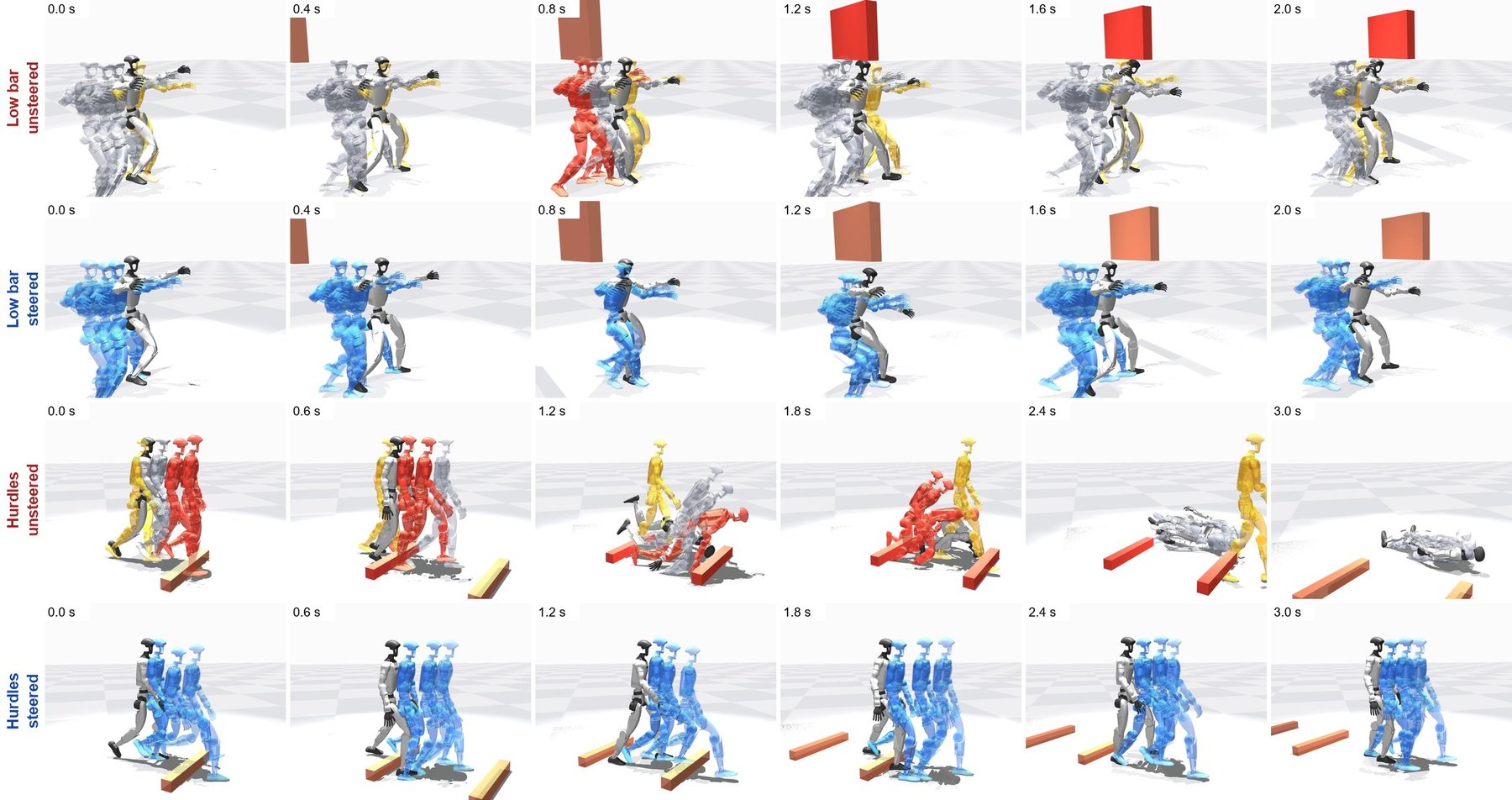}
  \caption{\textbf{Steering over time in simulation} (same frozen policy and reference in each pair of rows). Colors as in \cref{fig:ghost}: the executed robot in grey, the plan as ghosts (red where it hits the obstacle, blue after steering), the reference in gold. Unsteered, the plan runs into the low bar and over the hurdles, and the robot falls at the hurdles; steered, the plan lowers the body under the bar and lifts the feet over each hurdle, and the robot follows it.}
  \label{fig:ghost_seq}
\end{figure*}

\begin{figure*}[t]
  \centering
  \includegraphics[width=0.9\textwidth]{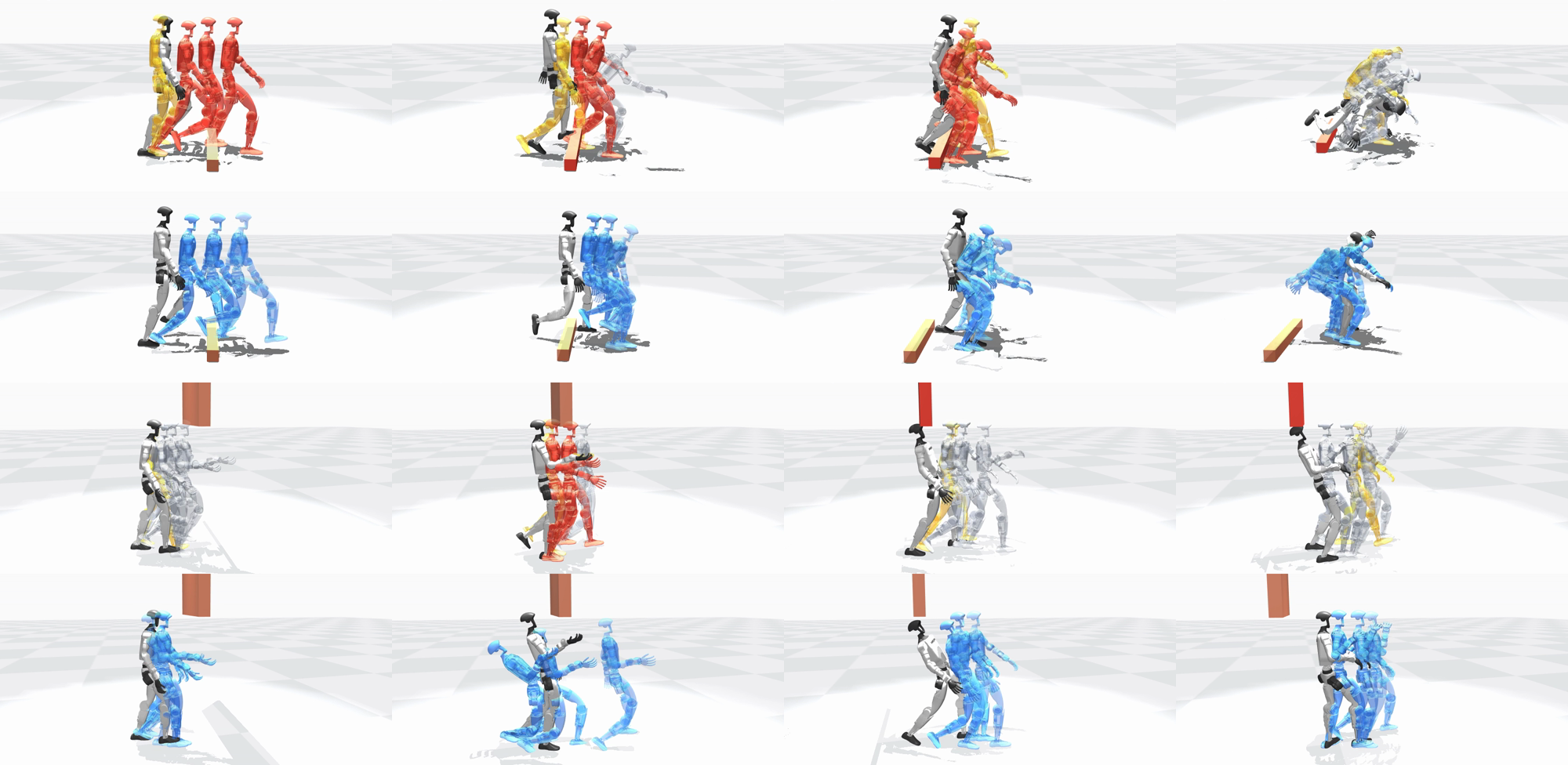}
  \caption{\textbf{Plan ghosts before and after steering.} Four time steps of one rollout each (left to right), for a hurdle (top two rows) and a low bar (bottom two rows). For each scene, the upper row is unsteered: grey ghosts show the 0.96\,s plan, red marks plan poses that hit the obstacle, and gold is the reference motion. The lower row is the same reference after test-time steering (blue ghosts).}
  \label{fig:ghost}
\end{figure*}

\paragraph{Steering moves the plan.}
\label{sec:app_ghost}
\cref{fig:ghost} shows the 0.96\,s plan of the generator before and after steering, for a hurdle and for a 1.25\,m bar.
Without steering, the planned poses follow the reference and overlap the obstacle (red).
With the cost, the generator returns a different plan (blue) that clears the obstacle, and the robot executes this plan.
The reference, the weights and the executor are unchanged; only the input noise moves.
\cref{fig:ghost_seq} shows the same over a longer stretch for two other motions, a low bar while carrying a heavy object and a row of hurdles while walking in a loop.

\begin{figure*}[t]
  \centering
  \setlength{\tabcolsep}{0.6pt}
  \begin{tabular}{ccccc}
  \includegraphics[width=0.194\textwidth]{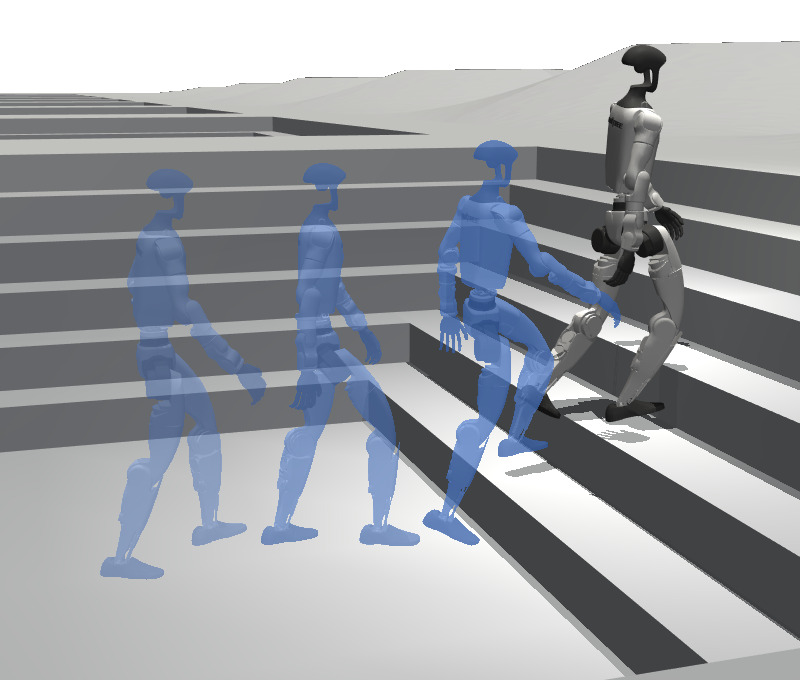} &
  \includegraphics[width=0.194\textwidth]{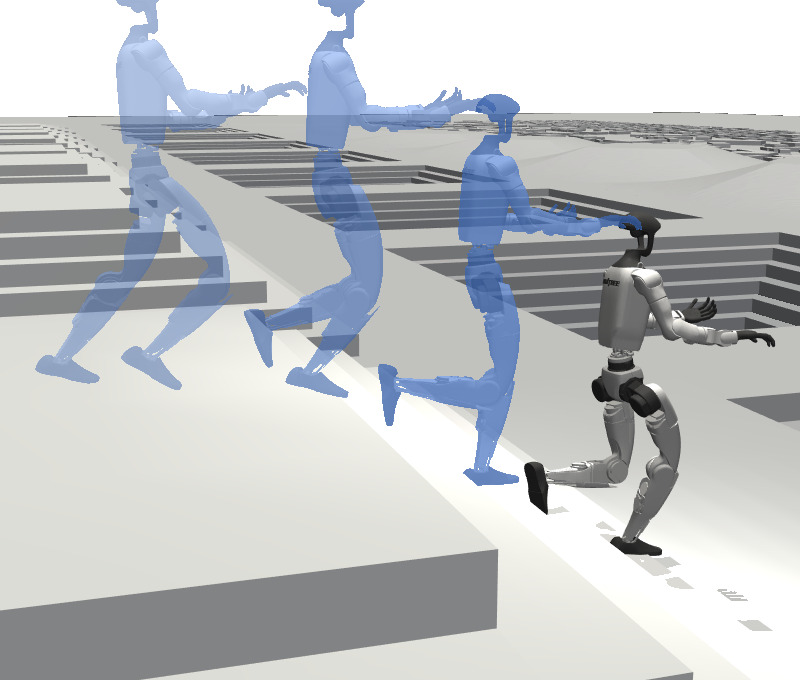} &
  \includegraphics[width=0.194\textwidth]{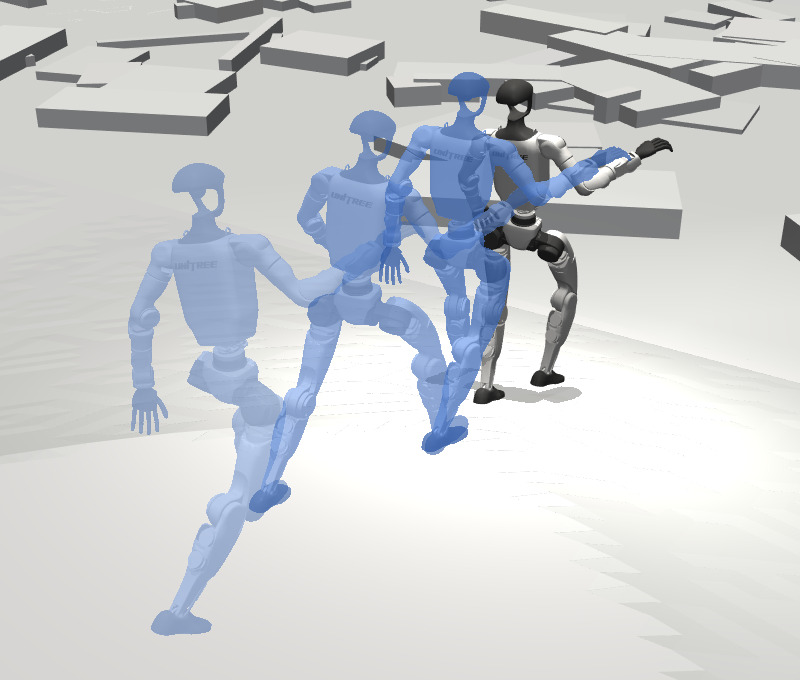} &
  \includegraphics[width=0.194\textwidth]{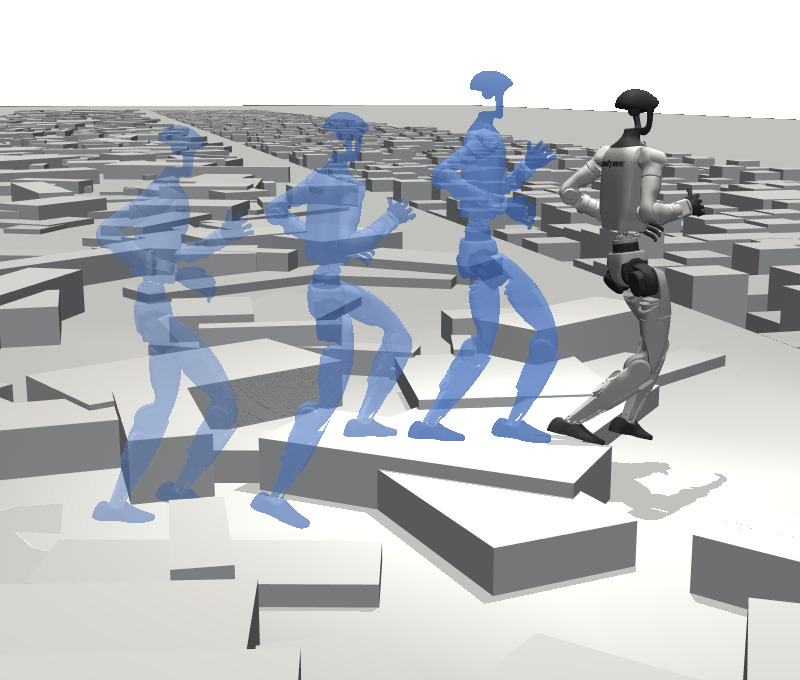} &
  \includegraphics[width=0.194\textwidth]{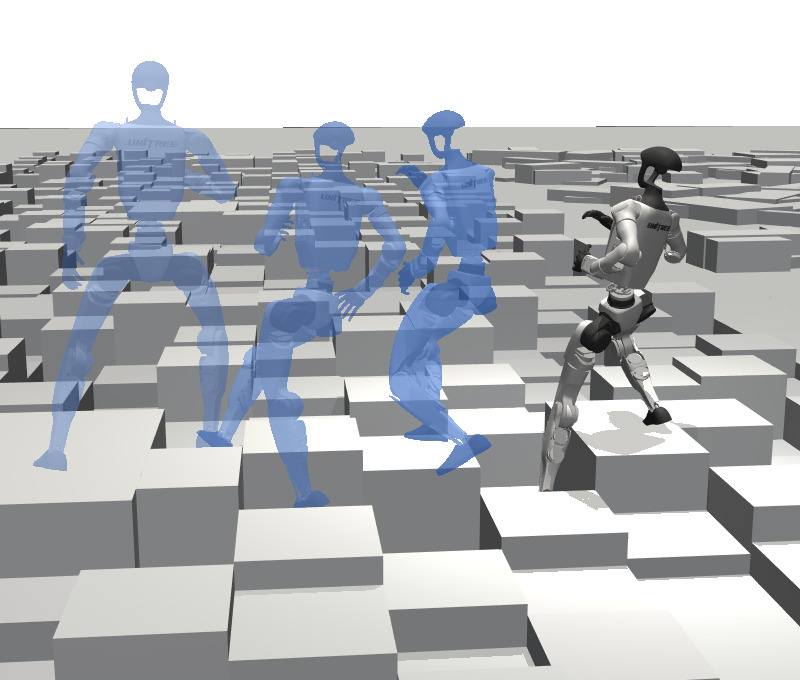} \\
  \small Stairs up (level 8) & \small Stairs down (level 8) & \small Slope (level 6) & \small Boxes (level 9) & \small Rough ground (level 7)
  \end{tabular}
  \caption{\textbf{One terrain-conditioned model on five terrain families} (simulation). Each panel overlays four poses, about 0.55\,m apart, of one rollout (earlier ones in blue); the reference is on flat ground, and the policy sees the terrain only through the height-scan tokens.}
  \label{fig:terrain_main}
\end{figure*}

\paragraph{Adding a condition: terrain.}
\label{sec:terrain}
Starting from the model trained from scratch (\cref{tab:track}), we encode a $21{\times}21$ height scan around the pelvis into 36 tokens that are appended to the input sequence of the generator; the reference stays a flat-ground motion and the inverse-dynamics head is unchanged.
After RL fine-tuning on stairs, slopes, boxes and rough ground, the terrain-conditioned \method{} succeeds on 0.90 of the cells (\cref{fig:terrain_main}) against 0.43 for the same model without the scan, and masking the tokens of the fine-tuned weights gives 0.55, so the gain comes from the map (\cref{tab:terrain}).
On flat ground, success is unchanged and the local error slightly higher (success 0.963 vs.\ 0.963, $E_l$ 27.6 vs.\ 26.0\,mm, one run).
The terrain-conditioned model remains steerable, with about half the gain of the model without terrain input, and has not been run on the real robot.

\begin{table}[t]
\centering
\small
\setlength{\tabcolsep}{3pt}
\resizebox{\linewidth}{!}{%
\begin{tabular}{@{}lccccccc@{}}
\toprule
Success$\uparrow$ & Flat & Down & Up & Slope & Boxes & Rough & All \\
\midrule
No terrain input & 0.87 & 0.53 & 0.16 & 0.55 & 0.19 & 0.32 & 0.43 \\
Terrain tokens masked & 0.90 & 0.81 & 0.25 & 0.67 & 0.42 & 0.37 & 0.55 \\
Terrain tokens (ours) & \textbf{0.97} & \textbf{0.99} & \textbf{0.91} & \textbf{1.00} & \textbf{0.86} & \textbf{0.76} & \textbf{0.90} \\
\bottomrule
\end{tabular}}
\caption{\textbf{Terrain as an extra input.} Success (no fall, ends within 1\,m of the reference end point) of 88 clips crossing six terrain families, mean of four evaluation passes in simulation (standard deviation of the last column 0.06, 0.06 and 0.02). ``Masked'' keeps the fine-tuned weights and masks the tokens. Down and Up are stairs; the reference is a flat-ground motion.}
\label{tab:terrain}
\end{table}

\paragraph{Terrain conditioning.}
\label{sec:app_terrain}
The terrain-conditioned model adds one input to the generator and leaves the rest unchanged.
The scan is $21{\times}21$ points at 0.1\,m spacing over $[-1,1]$\,m around the pelvis, in the pelvis yaw frame; each value is the pelvis height minus the ground height.
A small CNN maps it to a $6{\times}6$ grid, and each cell becomes one token (with a position embedding) at the end of the generator's input sequence, where it joins the other tokens by ordinary self-attention.
The attention of the existing tokens onto the terrain tokens carries a learned bias per layer, initialised very negative, so training starts from the unchanged tracker and the new tokens are opened by training.
The reference is a flat-ground motion that is re-anchored to the patch where the robot starts.
Training is RL through the learned world model, on a procedural terrain with six families (flat, stairs down, stairs up, slope, boxes, rough ground) and ten difficulty levels with a curriculum, and runs in two stages: RL on the terrain curriculum (several runs; the last, 30{,}000 iterations, took 7.7 hours on four GPUs), then 10{,}000 further iterations with references that conform to the terrain.
We evaluate with 88 clips that cross the terrain (a path longer than 2.5\,m), on cells of terrain family and level, and a clip counts as a success if the robot does not fall and ends within 1\,m of the end of the reference path.
Single evaluations vary by about $\pm$4 points, so \cref{tab:terrain} averages four passes.
\Cref{fig:terrain_main} shows rollouts, and \cref{tab:terrain_flat} the cost on flat ground.

Two points qualify the claim.
First, terrain-conditioned \method{} remains steerable: on the hurdle windows, steering raises its success from 0.65 to 0.76, about half the gain of the model without terrain input (0.32 to 0.55; \cref{tab:terrain_flat}).
Second, the model was evaluated only in simulation; the height-scan pipeline for the robot (a LiDAR-inertial map sampled on the same grid) is implemented but has not been run on hardware.

\begin{table}[t]
\centering
\small
\setlength{\tabcolsep}{3pt}
\resizebox{\linewidth}{!}{%
\begin{tabular}{@{}lcccc@{}}
\toprule
 & \multicolumn{2}{c}{Flat tracking} & \multicolumn{2}{c}{Hurdles, success$\uparrow$} \\
\cmidrule(lr){2-3}\cmidrule(lr){4-5}
Model & success$\uparrow$ & $E_l$\,(mm)$\downarrow$ & unsteered & steered \\
\midrule
No terrain input & 0.963 & 26.0 & 0.32 & 0.55 \\
Terrain-conditioned & 0.963 & 27.6 & 0.65 & 0.76 \\
\quad tokens off & 0.965 & 27.8 & 0.77 & 0.76 \\
\bottomrule
\end{tabular}}
\caption{\textbf{Cost of terrain training on flat ground} (training simulator, base: the model trained from scratch; flat tracking on the 512 BONES windows; hurdles: success, i.e.\ no fall, on the 79 walking windows, three runs for the steered column). The terrain-conditioned model sees a flat-ground height scan; ``tokens off'' runs its weights without the terrain tokens.}
\label{tab:terrain_flat}
\end{table}

\section{Design of steering}
\label{sec:steps}
\begin{table}[h]
\centering
\small
\setlength{\tabcolsep}{3.5pt}
\resizebox{\linewidth}{!}{%
\begin{tabular}{@{}lccc@{}}
\toprule
Variant (hurdles) & feasible$\uparrow$ & severe$\downarrow$ & succ.$\uparrow$ \\
\midrule
Both calls, 2 steps per call, $s{=}1$ & 0.68 & 0.18 & 0.91 \\
1 step per call & 0.51 & 0.29 & 0.75 \\
4 steps per call & 0.70 & 0.19 & 0.95 \\
6 steps per call + line search & \textbf{0.74} & \textbf{0.16} & 0.95 \\
Strength $s{=}0.5$ / $s{=}2$ & 0.56 / 0.66 & 0.23 / 0.19 & 0.87 / 0.91 \\
Best-of-8, no gradient & 0.38 & 0.42 & 0.58 \\
\bottomrule
\end{tabular}}
\caption{\textbf{Design of steering} (hurdles; both calls steered; an earlier batch than \cref{tab:steer_full}, whose re-run of the first row gives 0.66 / 0.19 / 0.92; three runs for the first row and for the line search, two for the other variants). Selecting among samples without gradient does not help; gradients do.}
\label{tab:steer_abl}
\end{table}

\paragraph{Steps and calls.}
\cref{tab:steer_abl} varies the steering procedure. One generator call is enough for tracking, but steering needs the second call: with one call, hurdles drop to 0.58 feasible (the level of editing the reference), the low bar at 1.25\,m to 0.76 clean, and reach to 0.43 (\cref{tab:calls}). A single gradient step is clearly worse (0.51), four steps change little (0.70), and a weaker strength hurts. Selecting among eight samples by cost without gradient steps (best-of-$K$) leaves the result at the unsteered level (0.38 vs.\ 0.40), so the gain comes from gradients on the generator input, not from sampling diversity.

\paragraph{Which call is steered.}
\cref{tab:calls} separates the calls (real-time strengths, two steps per steered call, three runs). Steering only the first call is largely undone by the second, which keeps $1-t_2=0.2$ of the steered draft and regenerates the rest: the low bar and reaching stay at the unsteered level, with small gains on hurdles (0.48 vs.\ 0.40) and the tray (ball lost 0.13 vs.\ 0.16). A single call with its step scaled by $t_2=0.8$, so that its length matches the second call's, gives the same as the full step (0.76 at 1.25\,m), so the gap to the second call (0.90) does not come from the step length. The input of the second call already carries a plan, and the generator's output responds to such an input more than to pure noise, which the training objective maps to nearly the same plan (\cref{sec:app_self}). Steering both calls is best on the tray (0.08 vs.\ 0.15 lost), where the two updates add up, and worse under the lower bars (0.84 vs.\ 0.90 at 1.25\,m).

\begin{table}[h]
\centering
\small
\setlength{\tabcolsep}{2pt}
\resizebox{\linewidth}{!}{%
\newcommand{\sd}[1]{{\scriptsize${\pm}$#1}}
\begin{tabular}{@{}lcccccc@{}}
\toprule
 & & \multicolumn{3}{c}{Low bar clean} & & Tray \\
\cmidrule(lr){3-5}
Steered call & Hurdles & 1.30\,m & 1.25\,m & 1.20\,m & Reach & lost$\downarrow$ \\
\midrule
None & 0.40 & 0.23 & 0.02 & 0.00 & 0.02 & 0.16 \\
First of two & 0.48\sd{0.01} & 0.32\sd{0.02} & 0.02\sd{0.00} & 0.00\sd{0.00} & 0.00\sd{0.00} & 0.13\sd{0.00} \\
Second of two (real-time) & 0.68\sd{0.04} & 0.95\sd{0.02} & \textbf{0.90}\sd{0.01} & \textbf{0.56}\sd{0.04} & \textbf{0.57}\sd{0.03} & 0.15\sd{0.01} \\
Both of two & 0.66\sd{0.03} & 0.95\sd{0.01} & 0.84\sd{0.02} & 0.52\sd{0.04} & 0.54\sd{0.06} & \textbf{0.08}\sd{0.01} \\
One call & 0.58\sd{0.02} & -- & 0.76\sd{0.04} & -- & 0.43\sd{0.02} & -- \\
One call, step $\times t_2$ & 0.60\sd{0.03} & 0.94\sd{0.01} & 0.76\sd{0.01} & 0.44\sd{0.03} & 0.42\sd{0.02} & 0.16\sd{0.01} \\
\bottomrule
\end{tabular}}
\caption{\textbf{Which generator call is steered} (79 walking windows, mean${\pm}$s.d.\ over three runs; unsteered one run). The single-call row with the full step is $N=1$ of \cref{tab:ncalls}. Tray: fraction of windows in which the ball leaves the tray (strength 0.5).}
\label{tab:calls}
\end{table}

\paragraph{Convergence.}
With two fixed-length steps the guidance does not converge for hard targets: for reaching and strengths of one and above, the final plan is farther from the target than the unguided plan. We test two procedures with a larger optimization budget, each applied unchanged to all three tasks (\cref{tab:steer_full}, three runs each): (A) 12 steps at 1/8 of the task's default strength (0.125 for hurdles, 0.5 for the low bar, 0.25 for reaching), and (B) 6 steps with a backtracking line search that halves a step up to four times (strength 1 for hurdles and reaching, 4 for the low bar). B is the best setting on hurdles (loose 0.74) and at the 1.30\,m bar (0.98 clean), ties the best at 1.20\,m (0.56), and raises the reach hit rate from 0.54 to 0.62; A raises it to 0.75, but its small steps do not clear the lower bars (0.22 clean at 1.25\,m). Neither procedure adds falls (success at least 0.95 on every task). Converging thus helps where two steps leave the cost far from zero (reaching, hurdles), while at the low bar the two-step settings are already as good (0.84--0.90 vs.\ 0.83 at 1.25\,m). The real-time setting is therefore a conservative operating point: where the latency budget allows, a larger budget raises the task results further with the same frozen model.

\paragraph{Executing guided plans.}
The action head executes guided plans without having seen them in training, and part of each edit is not realized. One reason is that the edit is not dynamically consistent: the one-step residual of a learned world model on the executed plans is 0.51 (normalized) for guided hurdle plans against 0.08 for unguided plans, mostly in root and joint velocities, since the edit moves joint angles without the matching velocities. A soft consistency cost in the guidance does not help: it shrinks the lift. Projecting the edited plan onto the world model (\cref{sec:app_method}) closes part of this gap: on hurdles, clean crossings rise from 0.19 to 0.32 and trailing-foot kicks fall from 0.42 to 0.29, with severe failures and success unchanged (three runs; the edit is smoothed over ticks with coefficient 0.75 in both settings). Joint-edit tasks improve as well: the realized reach error falls from 0.039 to 0.025\,m. For the low bar, which rewrites the root, the plain guided plan is already executed well (0.90 clean at 1.25\,m), and the optimized actions cannot replace the closed-loop action head (success 0.10 open loop).

\paragraph{Where the update acts.}
\label{sec:app_mech}
\cref{tab:mech} applies the same costs at different places, all in the real-time setting of \cref{tab:steer} (second call, two steps, default strengths and masks, three runs on the same 79 windows).
(i) A random direction with the same per-sample length as our update leaves every task at the unsteered level, so the gain does not come from perturbing the input.
(ii) Steering the plan itself, with gradient steps of the same normalized length on the generator's output and the result passed to the action head, matches our update at the 1.30\,m bar and is better at reaching (0.75 vs.\ 0.57). Reaching asks for one joint group to arrive at a target without a transition: a direct edit moves the wrist in the plan at once, and the head executes arm joints directly. Our update passes through the generator, which turns it into a smooth change of the whole plan, so it moves the wrist less per step and needs more steps: with 12 smaller steps it also hits 0.75 (\cref{tab:steer_full}). The direct edit is worse on hurdles (0.48 vs.\ 0.68) and under the lower bars (0.80 and 0.43 vs.\ 0.90 and 0.56), where the plan has to change consistently across legs, root and the horizon, which the generator provides and a direct edit does not.
(iii) Re-noising the pushed plan and regenerating it from the reference, as in SDEdit, erases the edit: every task falls to the unsteered level, since the generator maps the input back to a plan of the unchanged reference.
(iv) Giving the offline-edited reference of \cref{tab:steer} to \method{} as its condition, with the action head seeing it, is reference editing inside our own model; it is close to the teacher on the same edit (hurdles 0.46 vs.\ 0.58, 1.30\,m bar 0.64 vs.\ 0.72, reach 0.24 vs.\ 0.38) and far below steering, so the advantage of steering is not a weakness of the teacher as a tracker of edited references. That \method{} tracks the edit somewhat worse than the teacher is expected: it was distilled to imitate the teacher on natural references, an edited reference lies outside that distribution for both, and a student matches its teacher only where it was trained to.
(v) On the tray (strength 0.5, upper body hidden from the head), the direct edit fails outright: the ball leaves the tray in $0.97{\pm}0.01$ of the windows, against 0.16 unsteered and $0.15{\pm}0.01$ for our update in the same setting ($0.08{\pm}0.01$ with both calls steered, \cref{tab:calls}), and the local tracking error rises from 27.7 to 42.7\,mm. The hand tilt is reduced as much as with our update (11.6$^\circ$ vs.\ 11.8$^\circ$), so the ball is lost to the motion, not the tilt: a plausible reason is that the direct edit tilts the planned hand without the matching motion of the rest of the arm and body, and executing such a plan jolts the tray.
(vi) Sampling instead of the gradient: an MPPI update with eight perturbations per step, each as long as a gradient step and weighted by its cost, gains a little at the hurdles and the 1.30\,m bar (0.47 and 0.51) and nothing at the lower bars or reaching; sixteen perturbations do no better (hurdles 0.41, 1.25\,m bar 0.02, reach 0.03). Within two steps, the direction has to come from the gradient.
(vii) Limiting the size: projecting the steered input back into the ball of norm $r$ (in units of $\sqrt{Hd}$; pure noise has norm 1.0) after every step keeps the hurdles (0.68 and 0.66 at $r=1.5$ and 1.0) but lowers the 1.25\,m bar from 0.90 to 0.40 and 0.26 and reaching from 0.57 to 0.39 and 0.22. The projection scales the whole input, the draft included, so it bounds what a step near the training noise can do rather than isolating the effect of the norm.
Only the unconstrained gradient update on the generator input improves every task.

\begin{table}[h]
\centering
\small
\setlength{\tabcolsep}{2pt}
\resizebox{\linewidth}{!}{%
\newcommand{\sd}[1]{{\scriptsize${\pm}$#1}}
\begin{tabular}{@{}lccccc@{}}
\toprule
 & & \multicolumn{3}{c}{Low bar clean} & \\
\cmidrule(lr){3-5}
Where the cost acts & Hurdles & 1.30\,m & 1.25\,m & 1.20\,m & Reach \\
\midrule
Unsteered & 0.40 & 0.23 & 0.02 & 0.00 & 0.02 \\
Generator input, gradient (ours) & \textbf{0.68}\sd{0.04} & \textbf{0.95}\sd{0.02} & \textbf{0.90}\sd{0.01} & \textbf{0.56}\sd{0.04} & 0.57\sd{0.03} \\
Generator input, random direction & 0.41\sd{0.04} & 0.28\sd{0.02} & 0.02\sd{0.00} & 0.00\sd{0.00} & 0.04\sd{0.00} \\
Generator input, sampling (MPPI, 8) & 0.47\sd{0.02} & 0.51\sd{0.01} & 0.02\sd{0.00} & 0.00\sd{0.00} & 0.03\sd{0.01} \\
Generator input, gradient, $\|z\|\le 1.5$ & 0.68\sd{0.02} & 0.95\sd{0.02} & 0.40\sd{0.02} & 0.10\sd{0.03} & 0.39\sd{0.01} \\
Generator input, gradient, $\|z\|\le 1.0$ & 0.66\sd{0.02} & 0.89\sd{0.01} & 0.26\sd{0.02} & 0.05\sd{0.01} & 0.22\sd{0.01} \\
Plan, to the action head & 0.48\sd{0.01} & \textbf{0.95}\sd{0.01} & 0.80\sd{0.02} & 0.43\sd{0.02} & \textbf{0.75}\sd{0.02} \\
Plan, re-noised and regenerated & 0.40\sd{0.04} & 0.23\sd{0.01} & 0.02\sd{0.00} & 0.00\sd{0.00} & 0.01\sd{0.00} \\
Reference (offline edit), \method{} & 0.46\sd{0.05} & 0.64\sd{0.03} & 0.03\sd{0.00} & 0.00\sd{0.00} & 0.24\sd{0.02} \\
Reference (offline edit), teacher & 0.58\sd{0.02} & 0.72\sd{0.01} & 0.06\sd{0.00} & 0.00\sd{0.00} & 0.38\sd{0.01} \\
\bottomrule
\end{tabular}}
\caption{\textbf{Where the steering cost acts} (79 walking windows, real-time setting, mean${\pm}$s.d.\ over three runs; unsteered row and teacher row as in \cref{tab:steer_sd}). Hurdles: feasible passes; reach: hits within 8\,cm. Editing the plan takes about as long per control step as our update (9.7--13.9 vs.\ 10.6--13.7\,ms on one RTX 4090, by task), since the cost evaluation dominates both.}
\label{tab:mech}
\end{table}

\paragraph{From one step to many: a multi-step flow generator.}
Steering acts on the generator's input and its plan, not on the one-step sampler, so it covers generators from one step to many. To test this, we train the same model on the same data for the same 2{,}390 iterations with a time-conditioned flow-matching objective (logit-normal times, $x_0$ prediction), sample it with plain Euler steps, and steer the current Euler state before every step with the same costs, masks and strengths (\cref{tab:flow}). Unsteered, it tracks as well as \method{} (512 BONES windows: 95.9\% and 95.7\% success, $E_l$ 29.2 and 29.1\,mm with two and four Euler steps, against 95.3\% and 29.2\,mm for one \method{} call). With two Euler steps, i.e.\ two updates as in our real-time setting, it reaches the same task results within the spread of the runs except under the 1.25\,m bar (0.82 vs.\ 0.90); with four steps it clears more hurdles without contact (0.29), hits more wrist targets (0.73) and loses the ball on the tray less often (0.04 vs.\ 0.08 with both calls steered), but passes fewer low bars. At the same number of updates a steered step costs the same (timed on an otherwise idle RTX 4090, one clip, CUDA graph: 10.8--13.4\,ms with two guided Euler steps against 10.9--13.3\,ms for \method{} steering its second call, 17.9--21.1\,ms with four against 17.6--20.5\,ms with both calls steered).
The one-step generator differs in two ways. Nominal tracking, which the robot runs most of the time, takes one call (3.9\,ms) instead of two or four Euler steps (5.3 or 8.3\,ms). And the steered motion is smoother: in reaching, steering raises the jerk of the executed waist and arm joints from 400 to 915\,rad/s$^3$ with two Euler steps, against 381 to 571\,rad/s$^3$ for \method{} at the same number of updates, and the local tracking error by 7.1\,mm against 6.3\,mm. A plausible reason is that the last Euler steps act on a nearly clean plan and move the guided body part directly, as editing the plan does (\cref{tab:mech}); our update acts on the generator input, and the generator turns it into a plan that stays smooth and close to the reference.

\begin{table}[h]
\centering
\small
\setlength{\tabcolsep}{2pt}
\resizebox{\linewidth}{!}{%
\newcommand{\sd}[1]{{\scriptsize${\pm}$#1}}
\begin{tabular}{@{}lcccccccc@{}}
\toprule
 & & \multicolumn{2}{c}{Hurdles} & \multicolumn{3}{c}{Low bar clean} & & Upper-body \\
\cmidrule(lr){3-4}\cmidrule(lr){5-7}
Generator, steering & Updates & feasible & clean & 1.30\,m & 1.25\,m & 1.20\,m & Reach & jerk$\downarrow$ \\
\midrule
Flow, unsteered (2 Euler steps) & 0 & 0.46 & 0.02 & -- & 0.02 & -- & 0.01 & 400 \\
Flow, 2 Euler steps & 2 & 0.71\sd{0.01} & 0.14\sd{0.01} & 0.96\sd{0.01} & 0.82\sd{0.02} & 0.51\sd{0.02} & 0.60\sd{0.04} & 915 \\
Flow, 4 Euler steps & 4 & 0.67\sd{0.05} & 0.29\sd{0.02} & 0.94\sd{0.00} & 0.70\sd{0.07} & 0.44\sd{0.06} & 0.73\sd{0.03} & 963 \\
\midrule
\method{}, unsteered & 0 & 0.40 & 0.00 & 0.23 & 0.02 & 0.00 & 0.02 & 381 \\
\method{}, 2nd call, 2 steps & 2 & 0.68\sd{0.04} & 0.11 & 0.95\sd{0.02} & 0.90\sd{0.01} & 0.56\sd{0.04} & 0.57\sd{0.03} & 571 \\
\method{}, both calls, 4 steps & 4 & 0.66\sd{0.03} & 0.12 & 0.95\sd{0.01} & 0.84\sd{0.02} & 0.52\sd{0.04} & 0.54\sd{0.06} & 555 \\
\bottomrule
\end{tabular}}
\caption{\textbf{From one step to many: \method{} and a multi-step flow generator steered the same way} (same data and training length; 79 walking windows; mean${\pm}$s.d.\ over three runs, unsteered one run; \method{} rows as in \cref{tab:steer}). One update per Euler step. Hurdles: feasible passes may touch the hurdle with a foot, clean passes do not; reach: hits within 8\,cm; jerk: mean absolute third derivative of the executed waist and arm joint angles in reaching (rad/s$^3$; \method{} unsteered: the same windows without a target). Latencies are given in the text.}
\label{tab:flow}
\end{table}

\paragraph{Re-noise level and training length.}
\cref{tab:t2ck} varies the re-noise level $t_2$ of the second call and steers intermediate checkpoints, in the real-time setting. The 1.25\,m bar and reaching change by at most 0.05 for $t_2$ between 0.7 and 0.9, and tracking does not change; hurdles prefer a lower level, which keeps more of the draft (0.74 at $t_2=0.6$, 0.60 at 0.9). We keep $t_2=0.8$, which was fixed before this sweep. Steerability is present early in training: after 600 of the 2{,}390 iterations the model already passes 0.63 of the hurdles and 0.76 of the 1.25\,m bars.

\begin{table}[h]
\centering
\small
\setlength{\tabcolsep}{3pt}
\resizebox{\linewidth}{!}{%
\newcommand{\sd}[1]{{\scriptsize${\pm}$#1}}
\begin{tabular}{@{}llcccc@{}}
\toprule
 & & Hurdles & 1.25\,m bar & Reach & Tracking $E_l$ (mm) \\
\midrule
\multirow{4}{*}{$t_2$} & 0.6 & 0.74\sd{0.03} & 0.85\sd{0.02} & 0.53\sd{0.01} & 28.9 \\
 & 0.7 & 0.73\sd{0.03} & 0.91\sd{0.02} & 0.57\sd{0.01} & 28.9 \\
 & 0.8 (used) & 0.68\sd{0.04} & 0.90\sd{0.01} & 0.57\sd{0.03} & 29.0 \\
 & 0.9 & 0.60\sd{0.04} & 0.86\sd{0.02} & 0.55\sd{0.02} & 29.0 \\
\midrule
\multirow{3}{*}{Iteration} & 600 & 0.63 & 0.76 & 0.58 & 28.7 \\
 & 1{,}200 & 0.74 & 0.74 & 0.52 & 29.0 \\
 & 1{,}800 & 0.69 & 0.92 & 0.63 & 28.9 \\
\bottomrule
\end{tabular}}
\caption{\textbf{Re-noise level and training length} (79 walking windows, real-time setting; $t_2$ rows mean${\pm}$s.d.\ over three runs, checkpoint rows one run each, without the weight averaging of the final model, which is the $t_2=0.8$ row). Tracking: 512 BONES windows of the training simulator.}
\label{tab:t2ck}
\end{table}

\paragraph{Size of the input update.}
Normalized by $\sqrt{Hd}$, a pure-noise input has norm 1.0 and the re-noised input of the second call 0.82. The update moves only the samples whose plan violates the cost (15--64\% of the samples, by task) and, within a sample, only the entries $M_\text{guide}$ selects. In the real-time setting, moved inputs end with a mean norm of 1.2 (hurdles), 2.2 (reaching) and 2.4 (1.25\,m bar; median 3.2), and the displacement of the selected entries is 1.8, 5.1 and 3.9 times their own norm. The steered input thus leaves the shell of the training noise by a wide margin, and its plans are not dynamically consistent (world-model residual 0.51 against 0.08 unsteered, \cref{sec:steps}); yet the action head executes them (success at least 0.95, \cref{tab:steer_full}), and a random update of the same length does nothing (\cref{tab:mech}). The gain needs the direction of the cost gradient and, for the harder tasks, a step that leaves the shell of the training noise (\cref{tab:mech}, (vii)).

\paragraph{Held-out validation windows.}
All steering settings (strengths, masks, numbers of steps) were chosen on the 79 windows of \cref{tab:steer}. We draw 79 other walking windows, one per take, from held-out BONES takes that share no take with those 79 and no source motion with training, and rerun the unchanged settings (\cref{tab:val}). Every steered number is within 0.06 of its value on the original windows or above it, and the real-time setting is within 0.02 or above on every task; reaching improves (0.62 vs.\ 0.57). The settings therefore did not overfit the windows they were chosen on. The teacher baselines, rerun on the validation windows with the same editors, keep the ordering on obstacles (hurdles at most 0.47 and the 1.25\,m bar at most 0.14, against 0.68 and 0.92); in reaching, closed-loop editing stays ahead of the real-time setting (0.78 vs.\ 0.62), as on the original windows (0.60 vs.\ 0.57).

\begin{table}[h]
\centering
\small
\setlength{\tabcolsep}{2pt}
\resizebox{\linewidth}{!}{%
\newcommand{\sd}[1]{{\scriptsize${\pm}$#1}}
\begin{tabular}{@{}llccccc@{}}
\toprule
 & & & \multicolumn{3}{c}{Low bar clean} & \\
\cmidrule(lr){4-6}
Windows & Method & Hurdles & 1.30\,m & 1.25\,m & 1.20\,m & Reach \\
\midrule
Original & teacher, original ref. & 0.48 & 0.36 & 0.02 & 0.00 & 0.01 \\
 & teacher, offline edit & 0.58 & 0.72 & 0.06 & 0.00 & 0.38 \\
 & teacher, closed-loop edit & 0.57 & 0.87 & 0.11 & 0.02 & 0.60 \\
 & \method{}, unsteered & 0.40 & 0.23 & 0.02 & 0.00 & 0.02 \\
 & \method{}, 2nd call & 0.68\sd{0.04} & 0.95\sd{0.02} & 0.90\sd{0.01} & 0.56\sd{0.04} & 0.57\sd{0.03} \\
 & \method{}, both calls & 0.66\sd{0.03} & 0.95\sd{0.01} & 0.84\sd{0.02} & 0.52\sd{0.04} & 0.54\sd{0.06} \\
\midrule
Validation & teacher, original ref. & 0.47\sd{0.02} & 0.40\sd{0.01} & 0.02\sd{0.00} & 0.00\sd{0.00} & 0.03\sd{0.00} \\
 & teacher, offline edit & 0.42\sd{0.03} & 0.67\sd{0.00} & 0.14\sd{0.01} & 0.00\sd{0.00} & 0.37\sd{0.01} \\
 & teacher, closed-loop edit & 0.47\sd{0.02} & 0.80\sd{0.01} & 0.14\sd{0.00} & 0.00\sd{0.00} & 0.78\sd{0.01} \\
 & \method{}, unsteered & 0.41 & 0.31 & 0.00 & 0.00 & 0.03 \\
 & \method{}, 2nd call & 0.68\sd{0.02} & 0.96\sd{0.01} & 0.92\sd{0.01} & 0.56\sd{0.02} & 0.62\sd{0.02} \\
 & \method{}, both calls & 0.68\sd{0.01} & 0.94\sd{0.01} & 0.78\sd{0.03} & 0.50\sd{0.02} & 0.61\sd{0.04} \\
\bottomrule
\end{tabular}}
\caption{\textbf{Steering on held-out validation windows} with the settings chosen on the original 79 windows (79 windows each; steered rows mean${\pm}$s.d.\ over three runs, unsteered rows one run; teacher rows on the original windows as in \cref{tab:steer_sd}, on the validation windows three runs with the editors and settings of \cref{tab:steer}).}
\label{tab:val}
\end{table}

\paragraph{More generator calls.}
\method{} can also be sampled with more calls, re-noising the plan before call $k = 1, \dots, N{-}1$ to the descending level $t_k = 0.8\,(N-k)/(N-1)$ and steering every call with two steps (\cref{tab:ncalls}; $N=2$, $t_1=0.8$ is the both-calls row of \cref{tab:steer}). Tracking does not improve beyond one call (95.3--95.5\% success and 28.9--29.2\,mm on 512 BONES windows for every $N$), consistent with a generator trained for one step. In the harness of \cref{tab:track}, the exported model with one and with two calls gives 96.0\% and 96.0\% success on AMASS and 94.9\% and 94.9\% on BONES ($E_l$ 31.9 and 32.0, 27.2 and 27.3\,mm; the two-call row of \cref{tab:track} is an earlier export, whose AMASS numbers differ by the noise drawn inside the exported graph). For steering, four calls raise hurdles to 0.73 and reaching to 0.78, but lower clean passage under the 1.25\,m bar to 0.55, and eight calls are no better than four. A plausible reason is that the later calls start from little noise and return the plan toward the reference, which costs most where the plan must deviate most (the bar asks for a crouch the reference does not contain). Each call adds its forward and backward passes, so four steered calls take 35\,ms and miss the 50\,Hz budget; two calls remain the real-time choice.

\begin{table}[h]
\centering
\small
\setlength{\tabcolsep}{2pt}
\resizebox{\linewidth}{!}{%
\newcommand{\sd}[1]{{\scriptsize${\pm}$#1}}
\begin{tabular}{@{}cccccccc@{}}
\toprule
 & \multicolumn{2}{c}{Tracking (512)} & \multicolumn{3}{c}{Steering, all calls (79)} & \multicolumn{2}{c}{ms per step} \\
\cmidrule(lr){2-3}\cmidrule(lr){4-6}\cmidrule(l){7-8}
$N$ & Succ. & $E_l$ (mm) & Hurdles & 1.25\,m & Reach & Track & Steer \\
\midrule
1 & 95.3 & 29.2 & 0.58\sd{0.02} & 0.76\sd{0.04} & 0.43\sd{0.02} & 3.9 & 13.0 \\
2 & 95.5 & 29.0 & 0.68\sd{0.01} & \textbf{0.84}\sd{0.02} & 0.54\sd{0.06} & 5.2 & 20.5 \\
4 & 95.5 & 28.9 & \textbf{0.73}\sd{0.01} & 0.55\sd{0.04} & \textbf{0.78}\sd{0.01} & 8.2 & 35.3 \\
8 & 95.3 & 29.0 & 0.70\sd{0.01} & 0.45\sd{0.03} & 0.72\sd{0.01} & 13.9 & 64.9 \\
\bottomrule
\end{tabular}}
\caption{\textbf{Number of generator calls} (descending re-noise levels; every call steered with two steps; mean${\pm}$s.d.\ over three runs). Tracking on the 512 BONES windows of the training simulator. Times: RTX 4090, batch of one, the step captured as one CUDA graph; steering is timed on the 1.25\,m bar.}
\label{tab:ncalls}
\end{table}

\paragraph{Held-out dance.}
\label{sec:app_dance}
Steering was tuned on walking. We test it unchanged on dance: 40 windows of held-out BONES dance takes (no source motion shared with training; 31 from one performer, 9 from three others; styles such as krakowiak, charleston, heel-and-toe and slides), with the obstacle placed by a fixed rule at the frame of fastest root motion that is neither a turn (yaw change below 60$^\circ$ within 1\,s) nor a crouch (root above 0.70\,m), and windows kept only if the root moves at least 0.3\,m/s there. Each window is run three times with independent generator noise, with the real-time setting (second call, two steps, strength 4) and no per-window tuning. Against a head-height beam (1.18\,m) arriving frontally at 2 or 3\,m/s or sweeping from the side, clean passage rises from 0.01 to $0.72{\pm}0.03$ (0.72--0.73 for each variant) with falls at $0.02{\pm}0.01$, and the pinned arms deviate from the obstacle-free run by 4.0$^\circ$ in the obstacle window against 2.4$^\circ$ outside it. Over a static floor hurdle of 6 or 10\,cm, clean crossings rise from 0.00 to $0.36{\pm}0.06$ (0.40 and 0.33), with falls at the unsteered level (0.21 vs.\ 0.23), so steering turns contacts into clean crossings. Unlike in walking, where the hurdle meets a regular gait, the rule places the hurdle without regard to the step, so the foot reaches it at an arbitrary phase of a fast dance step.

\section{Self-assessment}
\label{sec:app_self}
\begin{table}[t]
\centering
\small
\setlength{\tabcolsep}{3pt}
\resizebox{\linewidth}{!}{%
\begin{tabular}{@{}llccc@{}}
\toprule
Score & Needs & AUC$\uparrow$ & Prec.\ top 2\%$\uparrow$ & Prec.\,/\,rec.\ top 5\%$\uparrow$ \\
\midrule
Noise spread (ours) & generator & 0.83 & \underline{0.91} & \underline{0.68} / \underline{0.50} \\
Plan deviation & generator & 0.82 & 0.68 & 0.44 / 0.33 \\
Kinematic rarity (kNN) & library & 0.81 & 0.23 & 0.32 / 0.23 \\
Critic value & RL critic & \underline{0.86} & 0.85 & 0.57 / 0.42 \\
Imagined rollout & policy, world model & \textbf{0.91} & \textbf{0.93} & \textbf{0.71} / \textbf{0.52} \\
Random & -- & 0.50 & 0.07 & 0.07 / 0.05 \\
\bottomrule
\end{tabular}}
\caption{\textbf{Predicting tracking failures without simulation} (scratch model, 7{,}884 clips, 537 failures; scores on the first 10\,s of each clip, on whole clips the spread has AUC 0.84 and top-2\% precision 0.90, \cref{sec:app_cost}). Rarity: kNN distance of a clip's reference statistics to the library. Imagined rollout: the policy is rolled out in its learned world model instead of the simulator, which needs an RL-trained world model and costs one policy and one world-model step per tick.}
\label{tab:detect}
\end{table}

\paragraph{Without a world model.}
The scores of \cref{tab:detect} are computed on the model trained from scratch, which also has a world model and a critic. The spread needs neither, so we compute it on the distilled model, which has none, with the same protocol, and label failures by one simulated rollout of the distilled model per clip (666 of the 7{,}884 clips fail, 8.4\%). The spread reaches AUC 0.80 and a top-2\% precision of 0.86 (top 5\%: 0.67 precision, 0.39 recall; whole clips: 0.81 and 0.87), against 0.83 and 0.91 on the scratch model; the plan deviation has a higher AUC (0.86) but a lower top-2\% precision (0.69). Failure discovery by the spread therefore carries over to the distilled controller that steers and runs on the robot.

\paragraph{Why the spread measures model error.}
Fix a condition $c$ (proprioception and reference tokens) and let $X^0\in\mathbb R^{D}$ be the future that is realized from it; the plan has $D=Hd=24\cdot102=2448$ entries, each standardized over the training data. At inference the generator is called once on pure noise, $z=\epsilon$, which is independent of $X^0$ given $c$. Writing $\bar f(c)=\mathbb E_\epsilon f_\theta(\epsilon,c)$, the expected loss of this call splits exactly into three terms:
\begin{equation}
\begin{split}
\mathbb E_{\epsilon,X^0}\big\|f_\theta(\epsilon,c)-X^0\big\|^2
&=\underbrace{\operatorname{tr}\operatorname{Cov}(X^0\mid c)}_{\text{irreducible}}
+\underbrace{\big\|\bar f(c)-\mathbb E[X^0\mid c]\big\|^2}_{\text{bias}}\\
&\quad+\underbrace{\mathbb E_\epsilon\big\|f_\theta(\epsilon,c)-\bar f(c)\big\|^2}_{\text{spread}},
\end{split}
\end{equation}
because $\epsilon$ and $X^0$ are independent given $c$, so the cross terms vanish. The irreducible term, \ie the genuine multimodality of the future, does not enter the spread; the diversity of a multi-step sampler, in contrast, is mostly that term. The constant predictor $\mathbb E[X^0\mid c]$ attains the irreducible term alone, so bias and spread are the excess of $f_\theta$ over it on pure-noise inputs, and the spread, which needs neither $X^0$ nor a simulator, is a lower bound on that excess.

\paragraph{The trained optimum is nearly constant on pure noise.}
Training does not use pure noise only: the generator predicts $X^0$ from $Y=(1-T)X^0+T\epsilon$ with $T=\operatorname{sigmoid}(U)$, $U\sim\mathcal N(0.8,0.8^2)$, and it does not see $T$. Its loss-optimal form is therefore
\begin{equation*}
\begin{split}f^\ast(y,c)&=\mathbb E[X^0\mid Y{=}y,c]\\&=\int\mathbb E[X^0\mid Y{=}y,T{=}t,c]\;p(t\mid y,c)\,dt,\end{split}
\end{equation*}
which in general depends on $y$, and the step from ``$\epsilon$ is independent of $X^0$'' to ``the optimum is constant in $\epsilon$'' needs two further facts. We show them for a Gaussian approximation $X^0\mid c\sim\mathcal N(\mu,\Sigma)$. The channels are standardized over the data, so $\operatorname{tr}\Sigma\le D$, but the largest eigenvalue $\lambda_{\max}$ of $\Sigma$ may exceed 1, since the channels of a plan are correlated. We write $\delta=1-t$ and $m=\|\mu\|^2/D\le 1$.

\emph{(a) The noise level is evident from the input.} Given $t$, $Y\sim\mathcal N\big((1-t)\mu,\;t^2I+\delta^2\Sigma\big)$. A pure-noise input has $\|\epsilon\|^2=D\,(1+O(D^{-1/2}))$ and $\langle\epsilon,\mu\rangle=O(\sqrt{mD})$, so
\begin{equation*}
\begin{split}\log p(\epsilon\mid t,c)={}&\mathrm{const}-D\,\delta^2(1+m/2)\\&+O(D\delta^3+\sqrt{mD}\,\delta):\end{split}
\end{equation*}
the linear terms in $\delta$ cancel, the terms in $\Sigma$ cancel at order $\delta^2$ for a typical $\epsilon$ (since $\mathbb E\,\epsilon^\top\Sigma\epsilon=\operatorname{tr}\Sigma$) and leave $O(\delta^3\operatorname{tr}\Sigma)$, and every level away from pure noise is penalized by about $D\delta^2$ nats (24 at $\delta=0.1$, 600 at $\delta=0.5$). The training prior puts only 0.4\% of its mass on $\delta<0.05$, but with this likelihood the posterior $p(t\mid\epsilon,c)$ has its mode at $\delta=0.022$--$0.026$, a root-mean-square $\delta$ below 0.03, and 99\% of its mass at $\delta<0.05$ (numerical integration over the prior, $m\in[0,1]$). What matters is this posterior, not how rarely training draws $t\approx1$: in $D=2448$ dimensions a pure-noise input is read as almost pure noise. This is the reason denoising generators train well without noise conditioning~\cite{sun2025noise}.

\emph{(b) At these levels the optimum barely reads its input.} For the Gaussian, $\mathbb E[X^0\mid y,t,c]=\mu+\delta\,\Sigma\,(\delta^2\Sigma+t^2I)^{-1}(y-\delta\mu)$, whose Jacobian in $y$ is bounded by $(\delta/t^2)\,\Sigma$. The posterior depends on $\epsilon$ only through $\|\epsilon\|^2$ and $\langle\epsilon,\mu\rangle$, which concentrate, so to leading order
\begin{equation*}
\begin{split}&\mathbb E_\epsilon\big\|f^\ast(\epsilon,c)-\mathbb E_\epsilon f^\ast(\epsilon,c)\big\|^2\\&\quad\approx\mathbb E\big[\delta^2/t^4\,\big|\,\epsilon\big]\operatorname{tr}\Sigma^2\le10^{-3}\lambda_{\max}\operatorname{tr}\Sigma ,\end{split}
\end{equation*}
that is, $10^{-3}\lambda_{\max}$ times the irreducible term. This floor is small unless the variation of the future given $c$ is concentrated in very few directions with $\lambda_{\max}$ in the hundreds; the irreducible term itself is then also large, which the spread does not measure. Under this approximation the loss-optimal generator thus maps every pure-noise input to nearly the same plan, its own spread is a negligible floor, and a spread above that floor is error of the trained model.

\paragraph{From the spread to the score.}
Our score is a robust, empirical version of the spread: $K{=}8$ samples, joint positions only, plan ticks from 4 on, and the 95th percentile over time. It is not itself the bound above, which concerns the squared error of the whole plan; it is a proxy motivated by it, and its relation to tracking failures is measured (\cref{tab:detect}), not derived.

Three limits apply. (i) The relation is one-sided: a small spread does not imply a small error, since the model can be wrong with little spread (bias); consistently, clips with low spread but rare motions still fail at 6.1\%, and rarity complements the spread. (ii) The floor of the optimum is derived for a Gaussian approximation of $X^0\mid c$; it gives the order of magnitude, not an exact value, and a multimodal future (\eg two ways around an obstacle) adds terms between modes that a Gaussian does not capture. (iii) The step from plan error to a tracking failure is empirical: the executed action is computed from the planned states, so a plan that varies with the noise yields actions that vary from tick to tick, but the link is measured (\cref{tab:detect}), not derived. After fine-tuning, the spread falls by only about 7\% while half of the failures are fixed, which suggests that much of the gain comes from the action head and the closed loop rather than from a more accurate plan.

\paragraph{Detection.}
\cref{tab:detect} lists all scores on the 7{,}884-clip scan. The top 329 clips by noise spread contain 242 failures (74\%), against 50\% for plan deviation and 34\% for kinematic rarity. The fine-tuning list below is taken after removing the held-out failure sets and evaluation clips from the ranking, so it contains fewer failures (187 of 329 failed in the scan; 154 of them are among the base model's 501 failures on the library rerun). Spread and rarity are complementary: clips above the median in both fail at 15.5\%, clips below it in both at 0.4\% (combined AUC 0.87). A threshold chosen on half of the clips (top ${\sim}6\%$) gives precision 0.61 and recall 0.49 on the other half.

\paragraph{Fine-tuning.}
\cref{tab:diff} gives the numbers behind \cref{fig:selfassess}b.
Each run starts from the scratch model (60k iterations) and trains for 8000 iterations with the same objective. Half of the environments draw the 329 clips selected by the score (clips whose names mark object interactions removed; obstacle jumps and step-overs are not marked and stay in the list, 40--44 clips per score, but the library rerun excludes them; held-out families excluded; these families are in the library the base model was trained on, and it fails on their held-out clips); the other half draw 16{,}000 library clips with a consolidation loss toward the frozen initial policy (weight 1) and keep their RL return (weight 0.3). The library rerun covers 7{,}788 of the scanned clips and is compared clip by clip with the base model, which fails on 501 of them.
Each selection is trained with three seeds: imagined rollout, noise spread, critic value and kinematic rarity keep their list, and the random control draws a new set of 329 clips uniformly from the feasible library for each run, so its three runs differ in their lists (two of them share training seed 0); this pool is not restricted to the scanned clips, so a random list holds only 0--3 of the base model's scanned failures. It is otherwise identical: same initialization, objective, replay of library clips, consolidation and number of iterations. Below, values are mean${\pm}$s.d.\ over the three seeds.
Failures outside a run's own 329-clip list are fixed at $12.2{\pm}4.8\%$ (base re-evaluated), $27.0{\pm}2.0\%$ (random), $42.1{\pm}4.1\%$ (imagined rollout), $41.2{\pm}3.3\%$ (critic value), $29.8{\pm}2.8\%$ (kinematic rarity) and $44.2{\pm}5.6\%$ (noise spread).
Two further controls use the candidate pool of the score-ranked lists (the scanned clips without object interactions and without held-out or evaluation clips; 7{,}615 clips, 360 of them failures of the base model's scan). Random lists drawn from this pool, one per seed, hold 10--15 of these failures and fix $35.1{\pm}5.9\%$ of the base failures ($34.3{\pm}5.6\%$ outside the list), still below the noise spread with non-overlapping bands; their unseen-family success is $35.8{\pm}8.2\%$. A list of 329 of the 360 failures found by one simulated rollout per clip, i.e.\ selection by the labels themselves, fixes $50.7{\pm}3.9\%$ ($45.1{\pm}4.7\%$ outside the list) and reaches $46.4{\pm}0.9\%$ on unseen and $46.8{\pm}4.8\%$ on seen families: the noise spread matches it on fixed and unseen-family clips without running the simulator, while the rollout list is better on seen families.
Selecting by the noise spread therefore fixes $1.8\times$ as many failures as random library clips at the same budget ($48.3{\pm}4.2\%$ against $26.9{\pm}2.0\%$; $1.4\times$ against random clips from the same candidate pool, below) and lowers the library failure rate from 6.4\% to $5.0{\pm}0.4\%$ (random: $6.2{\pm}0.1\%$), with regressions of $1.8{\pm}0.2\%$ (random: $1.6{\pm}0.2\%$).
Clips selected by imagined rollouts are fixed at a comparable rate ($46.3{\pm}2.8\%$), with fewer regressions ($1.3{\pm}0.1\%$), a similar library failure rate ($4.7{\pm}0.2\%$) and the same success on seen families ($39.3{\pm}0.9\%$ against $38.7{\pm}4.3\%$); the noise spread generalizes better to unseen families ($47.9{\pm}2.3\%$ against $43.0{\pm}1.0\%$) and needs neither the policy rollouts nor the learned world model.
Lists ranked by the RL critic and by kinematic rarity hold fewer of the base model's failures (122 and 80 of 329, against 154 for the noise spread) and fix fewer of them ($41.2{\pm}2.3\%$ and $30.5{\pm}3.1\%$); on unseen families they reach $37.9{\pm}2.8\%$ and $35.2{\pm}3.7\%$, close to random clips.
The RL return on mastered clips is what keeps regressions near the noise floor: with the consolidation loss alone (an earlier run on another 329-clip list), 4.6\% of tracked clips break and the library failure rate rises from 6.4\% to 8.2\%.

\begin{table}[t]
\centering
\small
\setlength{\tabcolsep}{3pt}
\newcommand{\sd}[1]{{\scriptsize${\pm}$#1}}
\resizebox{\linewidth}{!}{%
\begin{tabular}{@{}lcccccc@{}}
\toprule
Hard clips selected by & Fixed$\uparrow$ & Broken$\downarrow$ & Fail rate$\downarrow$ & Unseen fam.$\uparrow$ & Seen fam.$\uparrow$ & Hard $E_l\downarrow$ \\
\midrule
None (base re-evaluated) & 12.2\sd{4.8} & \textbf{0.8}\sd{0.3} & 6.4\sd{0.0} & 24.5\sd{2.6} & 9.7\sd{2.3} & \textbf{33.6} \\
Random (same budget) & 26.9\sd{2.0} & 1.6\sd{0.2} & 6.2\sd{0.1} & 34.8\sd{3.8} & 17.2\sd{6.5} & 34.2\sd{0.7} \\
Random, same candidate pool & 35.1\sd{5.9} & 1.3\sd{0.3} & 5.4\sd{0.4} & 35.8\sd{8.2} & 25.8\sd{4.3} & 33.9\sd{1.2} \\
Imagined rollout & 46.3\sd{2.8} & 1.3\sd{0.1} & \textbf{4.7}\sd{0.2} & 43.0\sd{1.0} & 39.3\sd{0.9} & 35.1\sd{1.0} \\
Critic value & 41.2\sd{2.3} & 1.3\sd{0.1} & 5.0\sd{0.2} & 37.9\sd{2.8} & 31.7\sd{3.4} & 35.0\sd{0.8} \\
Kinematic rarity & 30.5\sd{3.1} & 1.3\sd{0.1} & 5.7\sd{0.3} & 35.2\sd{3.7} & 19.9\sd{0.9} & 34.0\sd{1.1} \\
Noise spread (ours) & 48.3\sd{4.2} & 1.8\sd{0.2} & 5.0\sd{0.4} & \textbf{47.9}\sd{2.3} & 38.7\sd{4.3} & 35.3\sd{0.4} \\
\midrule
Simulated rollout (labels) & \textbf{50.7}\sd{3.9} & 1.7\sd{0.5} & 4.8\sd{0.4} & 46.4\sd{0.9} & \textbf{46.8}\sd{4.8} & 34.1\sd{0.8} \\
\bottomrule
\end{tabular}}
\caption{\textbf{Targeted fine-tuning on 329 selected clips} (\%; mean${\pm}$s.d.\ over three training seeds, sample s.d.; None: the base model re-evaluated, two evaluations). Fixed: base-model failures on the 7{,}788-clip library that succeed after fine-tuning. Broken: tracked clips that fail after it. Fail rate: library failure rate after fine-tuning (base 6.4). Unseen / Seen fam.: success on 110 held-out failures of unseen skill families and on 62 of seen families. Hard $E_l$: local error (mm) on 40 hard clips the base model tracks. Same candidate pool: random lists drawn from the clips the scores rank. Simulated rollout: 329 of the 360 failures that one rollout per clip finds in that pool; these labels come from the same scan that defines Fixed, so its Fixed column is optimistic (outside its list: $45.1{\pm}4.7$).}
\label{tab:diff}
\end{table}

\paragraph{Selection cost.}
\label{sec:app_cost}
We time both scores on the same 1{,}000 random library clips (mean length 6.4\,s), with the same checkpoint, one worker each, back to back on the same RTX 4090, once on the whole clips and once on their first 10\,s. The noise spread ($K{=}8$, every fifth tick, all 100 clips of a chunk in one batch) takes 88\,s on the whole clips (86\,s of computation) and 71--91\,s on the first 10\,s (two runs). One simulated rollout per clip (mjlab, 100 clips per process in parallel, as in our scans) takes 489\,s on the whole clips and 451\,s on the first 10\,s; of this, the simulation loops take 180 and 147\,s and the rest is the start-up of each process (about 30\,s). On the whole clips the spread is therefore 2.4$\times$ cheaper than one rollout per clip if the simulator is started once, 5.6$\times$ as our scans ran, and 11--28$\times$ cheaper than the five rollouts per clip of Extreme-RGMT~\cite{xrgmt} (first 10\,s: 1.9--6.4$\times$ and 8--32$\times$). Five rollouts are used because one rollout is a noisy label: with another seed, 16--18\% of the clips that fail succeed. Ranking quality does not depend on the 10\,s cut: on the whole clips the spread has AUC 0.84 and precision 0.90 among the top 2\% (0.83 and 0.91 with the cut). The simulator numbers are for our mjlab stack, not for the simulator of Extreme-RGMT, and we compare cost only here; a list selected by one rollout per clip is trained in \cref{tab:diff}.